\documentclass[preprint,12pt]{elsarticle}

\usepackage{amssymb}
\usepackage{url}
\usepackage{subfigure}
\usepackage{amsmath}
\usepackage{threeparttable}
\usepackage{algorithm}
\usepackage{algorithmic}
\usepackage{epstopdf}
\usepackage{booktabs}
\usepackage{color}
\usepackage{multirow}

\newtheorem{prop}{Proposition}
\biboptions{numbers,sort&compress}

\usepackage{lineno}

\begin{document}

\begin{frontmatter}

\title{Two-dimensional Bhattacharyya bound linear discriminant analysis with its applications}



\author[1]{Yan-Ru Guo}
\address[1]{Department of Mathematics, Shanghai University, Shanghai 200444, P.R.China}

\author[1]{Yan-Qin Bai}

\author[2]{Chun-Na Li\corref{cor1}}
\cortext[cor1]{Corresponding author.}\ead{na1013na@163.com}
\address[2]{Management School, Hainan University, Haikou, 570228, P.R.China}

\author[3]{Lan Bai}
\address[3]{School of Mathematical Sciences, Inner Mongolia University, Hohhot, 010021, P.R.China}

\author[2]{Yuan-Hai Shao}


\address{}
\begin{abstract}
Recently proposed L2-norm linear discriminant analysis criterion via the Bhattacharyya error bound estimation (L2BLDA) is an effective improvement of linear discriminant analysis (LDA) for feature extraction. However, L2BLDA is only proposed to cope with vector input samples. When facing with two-dimensional (2D) inputs, such as images, it will lose some useful information, since it does not consider intrinsic structure of images. In this paper, we extend L2BLDA to a two-dimensional Bhattacharyya bound linear discriminant analysis (2DBLDA). 2DBLDA maximizes the matrix-based between-class distance which is measured by the weighted pairwise distances of class means and meanwhile minimizes the matrix-based within-class distance. The weighting constant between the between-class and within-class terms is determined by the involved data that makes the proposed 2DBLDA adaptive. In addition, the criterion of 2DBLDA is equivalent to optimizing an upper bound of the Bhattacharyya error. The construction of 2DBLDA makes it avoid the small sample size problem while also possess robustness, and can be solved through a simple standard eigenvalue decomposition problem. The experimental results on image recognition and face image reconstruction demonstrate the effectiveness of the proposed methods.
\end{abstract}
\begin{keyword}
feature extraction; dimensionality reduction; two-dimensional linear discriminant analysis; robust linear discriminant analysis; Bhattacharyya error bound
\end{keyword}

\end{frontmatter}


\section{Introduction}

Feature extraction plays an important role in pattern recognition. As a powerful supervised feature extraction method, linear discriminant analysis (LDA) \cite{Fisher36,Fukunaga90} has been successfully applied in many problems, such as face recognition \cite{BelhumeurHespanha97}, text mining \cite{AbuZeina18}, speech recognition \cite{Zeiler16}, and microarrays \cite{GuoHastie07,DongZhao16}.

However, classical LDA is a vector (or one-dimensional, 1D) based method. When the input data are naturally of matrix (or two-dimensional, 2D) form, such as images, it may lead to two issues. Firstly, converting 2D data to 1D data may produce high dimensional vector, and hence will suffer from the small sample size (SSS) problem. For example, a 32$\times 32$ face image corresponds to a 1024 dimensional vector. Secondly, during the transformation procedure from 2D data to 1D data, useful discriminant information may be lost \cite{LiPangYuan10}.
To handle this problem, Liu et al. suggested a 2D image matrix-based linear discriminant technique that performed LDA directly on images \cite{Liuetal93}, and two-dimensional LDA (2DLDA) was proposed and studied in \cite{LiYuan05,XiongSwamy05}. 2DLDA constructed the within-class scatter matrix and between-class scatter matrix by using the original image samples represented in matrix form rather than converting matrices to vectors beforehand. Compared to LDA, 2DLDA alleviated the SSS problem when some mild condition was satisfied \cite{Kong05,LiYuan05}, and can preserve the original structure of the input matrix.

Thereafter, various modifications and improvements of 2DLDA were studied by many researchers. Since 2DLDA was based on L2-norm, it was sensitive to noise and outliers. To improve the robustness of 2DLDA, robust replacements of L2-norm were studied, including L1-norm \cite{L12DLDA15,ChenL12DLDA15,TrL12DLDA17,LiShangShao19}, nuclear norm \cite{LuYuanLai18,ZhangDengNie19}, Lp-norm \cite{LiShaoWang19,G2DLDA}, and Schatten Lp-norm, $0<p<1$ \cite{Duetal17}. Some of the studies focused on extracting the discriminative transformations on both sides of the matrix samples.
The authors in \cite{YangZhang05,Kong05,Noushath06} implemented 2DLDA on matrices in sequence or independently, and then combined left and right sides transformations to achieve bilateral dimensionality reduction. Ye et al. \cite{YeJanardan05} and Li et al. \cite{LiShaoWang19} used iterative schemes to extract transformations on both sides.
Extensions to other machine learning problems and real applications were also investigated. For example, Wang et al. \cite{WangQinetal17} proposed a convolutional 2DLDA for nonlinear dimensionality reduction, and Xiao et al. \cite{XiaoChenGong19} studied a two-dimensional
quaternion sparse discriminant analysis that met the requirements of representing RGB and RGB-D images.

Though 2DLDA can ease the SSS problem, it may still face the singularity issue theoretically as LDA since it needs to solve a generalized eigenvalue problem.
Recently, a novel vector based L2-norm linear discriminant analysis criterion via the Bhattacharyya error bound estimation (L2BLDA) \cite{L2BLDA} was proposed.
Compared to LDA, L2BLDA solved a simple standard eigenvalue decomposition problem rather than a generalized eigenvalue decomposition problem, which avoided the singularity issue and had robustness.
In fact, minimizing the Bhattacharyya error \cite{Bhattacharyya43} bound is a reasonable way to establish classification \cite{Devijveretal82,Fukunaga90}.
In this paper, inspired by L2BLDA, to cope with the SSS problem and improve the robustness of 2DLDA, we first derive a Bhattacharyya error upper bound for matrix input classification, and then propose a novel two-dimensional linear discriminant analysis by minimizing this Bhattacharyya error upper bound, called 2DBLDA. The proposed 2DBLDA has the following characteristics:

$\bullet$ 2DBLDA criterion is derived by minimizing a close Bhattacharyya error bound, and therefore the rationality of 2DBLDA is guaranteed.

$\bullet$ The between-class distance of 2DBLDA is characterized by the sum of weighted distances between each pair of class means, which is benefit to balance different classes.

$\bullet$ The between-class distance and the within-class distance of 2DBLDA are weighted by a constant which is calculated according to the input data. This constant not only helps the objective of 2DBLDA achieve minimum error bound, but also makes the proposed 2DBLDA adaptive and without tuning any parameters. By combining the above weighted between-class distance, 2DBLDA improves the robustness of 2DLDA.

$\bullet$ 2DBLDA can be solved effectively through a standard eigenvalue decomposition problem, which does not involve the inverse of a matrix, and hence avoids the SSS problem.

$\bullet$ Experimental results on image recognition and face reconstruction demonstrate the effectiveness of 2DBLDA.

The paper is organized as follows. Section \ref{secRelated} briefly introduces LDA, L2BLDA and 2DLDA.
Section \ref{sec2DBLDA} proposes our 2DBLDA and gives the corresponding theoretical analysis. Section \ref{secExp} makes comparisons of 2DBLDA with its related approaches. At last, the concluding remarks are given in Section \ref{secCon}. The proof of the Bhattacharyya error upper bound of 2DBLDA is given in the Appendix.

The notations of this paper are given as follows. We consider a supervised learning problem in the $d_1\times d_2$-dimensional matrix space $\mathbb{R}^{d_1\times d_2}$. The training data set is given by $T=\{(\textbf{X}_{1},y_{1}),...,(\textbf{X}_{N},y_{N})\}$, where $\textbf{X}_{l}\in\mathbb{R}^{d_1\times d_2}$ is the $l$-th input matrix sample and $y_{l}\in \{1,...,c\}$ is the corresponding label, $l=1,...,N$. Assume that the $i$-th class contains $N_i$ samples, $i=1,\ldots,c$. Then we have $\sum\limits_{i=1}^{c}N_i=N$. We further write the samples in the $i$-th class as $\{\textbf{X}_{is}\}$, where $\textbf{X}_{is}$ is the $s$-th sample in the $i$-th class, $i=1,\ldots,c$, $s=1,\ldots,N_i$. Let $\overline{\textbf{X}}=\frac{1}{N}\sum\limits_{l=1}^{N}\textbf{X}_l$ be the mean of all matrix samples and ${\overline{\textbf{X}}}_i=\frac{1}{N_i}\sum\limits_{s=1}^{N_i}\textbf{X}_{is}$ be the mean of matrix samples in the $i$-th class. For a matrix $\textbf{Q}=(\textbf{q}_1,\,\textbf{q}_2,\ldots,\textbf{q}_n)\in \mathbb{R}^{m\times n}$, its  Frobenius norm (F-norm) $||\textbf{Q}||_F$ is defined as $||\textbf{Q}||_F=\sqrt{\sum\limits_{i=1}^{n}||\textbf{q}_i||_2^2}$. F-norm is a natural generalization of vector L2-norm to matrices.

\section{Related work}\label{secRelated}
\subsection{Linear discriminant analysis}\label{LDA}

Linear discriminant analysis (LDA) \cite{Fisher36,Fukunaga90} found a projection transformation matrix $\textbf{W}$ such that the ratio of between-class distance to within-class distance is maximized in the projected space. In specific, for data in $\mathbb{R}^n$, LDA found an optimal $\textbf{W}\in\mathbb{R}^{n\times r}$, $r\leq n$, such that the most discriminant information of the data is retained in $\mathbb{R}^r$ by solving the following problem
\begin{equation}\label{LDA}
\begin{split}
\underset{\textbf{W}}{\max}&~~\frac{\hbox{tr}(\textbf{W}^T\textbf{S}_b\textbf{W})}{\hbox{tr}(\textbf{W}^T\textbf{S}_w\textbf{W})},\\
\end{split}
\end{equation}
where $\hbox{tr}(\cdot)$ is the trace operation of a matrix, and the between-class scatter matrix $\textbf{S}_b$ and the within-class scatter matrix $\textbf{S}_w$ are defined by
\begin{equation}\label{SB}
\textbf{S}_b=\frac{1}{N}\sum\limits_{i=1}^{c}N_i({\overline{\textbf{x}}}_i-{\overline{\textbf{x}}})({\overline{\textbf{x}}}_i-{\overline{\textbf{x}}})^T
\end{equation}
and
\begin{equation}\label{SW} \textbf{S}_w=\frac{1}{N}\sum\limits_{i=1}^{c}\sum\limits_{s=1}^{N_i}(\textbf{x}_{is}-{\overline{\textbf{x}}_i})(\textbf{x}_{is}-{\overline{\textbf{x}}_i})^T,
\end{equation}
where $\overline{\textbf{x}}_{i}\in\mathbb{R}^{n}$ is the mean of the samples in the $i$-th class, $\overline{\textbf{x}}\in\mathbb{R}^{n}$ is the mean of the whole data, $\textbf{x}_{is}\in\mathbb{R}^{n}$ is the $s$-th sample of the $i$-th class. The optimization problem \eqref{LDA} is equivalent to the generalized problem $\textbf{S}_b\textbf{w}=\lambda \textbf{S}_w\textbf{w}$ where $\lambda\not=0$, with its solution $\textbf{W}=(\textbf{w}_1,\ldots,\textbf{w}_r)$ given by the first $r$ largest eigenvalues of $(\textbf{S}_w)^{-1}\textbf{S}_b$ in case $\textbf{S}_w$ is nonsingular.

\subsection{L2-norm linear discriminant analysis criterion via the Bhattacharyya error bound estimation}\label{secL2BLDA}

As an improvement of LDA, L2-norm linear discriminant analysis criterion via the Bhattacharyya error bound estimation (L2BLDA) \cite{L2BLDA} is a recently proposed vector-based weighted linear discriminant analysis. In the vector space $\mathbb{R}^{n}$, by minimizing an upper bound of the Bhattacharyya error, the optimization problem of L2BLDA is formulated as
\begin{equation}
\begin{split}\label{L2BLDA}
\underset{\textbf{W}}{\min}~~&-\frac{1}{N}\sum_{i<j}\sqrt{N_iN_j}||\textbf{W}^T(\overline{\textbf{x}}_{i}-\overline{\textbf{x}}_{j})||_2^2+\Delta\sum_{i=1}^{c}\sum_{s=1}^{N_i}||\textbf{W}^T(\textbf{x}_{is}-\overline{\textbf{x}}_i)||_2^2\\
\hbox{s.t.\ }& \textbf{W}^T\textbf{W}=\textbf{\textbf{I}},
\end{split}
\end{equation}
where $\textbf{W}\in\mathbb{R}^{n\times r}$, $r\leq n$, $P_i=\frac{N_i}{N}$, $P_j=\frac{N_j}{N}$, $\overline{\textbf{x}}_{i}\in\mathbb{R}^{n}$ is the mean of the samples in the $i$-th class, $\textbf{x}_{is}\in\mathbb{R}^{n}$ is the $s$-th sample of the $i$-th class,
$\Delta=\frac{1}{4}\sum\limits_{i<j}^c\sqrt{P_iP_j}||{\overline{\textbf{x}}_{i}-\overline{\textbf{x}}_{j}}||_2^2$,
and $\textbf{\textbf{I}}\in\mathbb{R}^{r\times r}$ is the identity matrix.


L2BLDA can be solved through the following standard eigenvalue decomposition problem
\begin{equation}
\begin{split}\label{L2BLDA2}
\underset{\textbf{W}}{\min}&~~\hbox{tr}(\textbf{W}^T\textbf{S}\textbf{W})\\
\hbox{s.t.\ }& \textbf{W}^T\textbf{W}=\textbf{I},
\end{split}
\end{equation}
where
\begin{equation}
\begin{split}
\textbf{S}=&-\frac{1}{N}\sum_{i<j}\sqrt{N_iN_j}(\overline{\textbf{x}}_{i}-\overline{\textbf{x}}_{j})(\overline{\textbf{x}}_{i}-\overline{\textbf{x}}_{j})^T+\Delta\sum_{i=1}^{c}\sum_{s=1}^{N_i}(\textbf{x}_{is}-\overline{\textbf{x}}_i)(\textbf{x}_{is}-\overline{\textbf{x}}_i)^T.
\end{split}
\end{equation}
Then $\textbf{W}=(\textbf{w}_1,\textbf{w}_2,\ldots,\textbf{w}_r)$ is obtained by the $r$ orthogonormal eigenvectors that correspond to the first $r$ nonzero smallest eigenvectors of $\textbf{S}$. After obtaining optimal $\textbf{W}$, a new sample $\textbf{x}\in\mathbb{R}^{n}$ is projected into $\mathbb{R}^{r}$ by $\textbf{W}^T\textbf{x}$.

\subsection{Two-dimensional linear discriminant analysis}\label{2DL2LDA}

Different from LDA or L2BLDA that works on vector samples, two-dimensional linear discriminant analysis (2DLDA) \cite{Kong05,LiYuan05} operates on matrix ones. 2DLDA defines the between-class scatter matrix and the within-class scatter matrix directly on the 2D data set $T$ as
\begin{equation}\label{Sb}
\textbf{S}_b=\frac{1}{N}\sum\limits_{i=1}^{c}N_i({\overline{\textbf{X}}}_i-{\overline{\textbf{X}}})({\overline{\textbf{X}}}_i-{\overline{\textbf{X}}})^T
\end{equation}
and
\begin{equation}\label{Sw} \textbf{S}_w=\frac{1}{N}\sum\limits_{i=1}^{c}\sum\limits_{s=1}^{N_i}(\textbf{X}_{is}-{\overline{\textbf{X}_i}})(\textbf{X}_{is}-{\overline{\textbf{X}_i}})^T.
\end{equation}
Then 2DLDA solves the following optimization problem
\begin{equation}\label{2DLDA}
\begin{split}
\underset{\textbf{W}}{\max}&~~\frac{\hbox{tr}(\textbf{W}^T\textbf{S}_b\textbf{W})}{\hbox{tr}(\textbf{W}^T\textbf{S}_w\textbf{W})}=\frac{\sum\limits_{i=1}^{c}N_i\|\textbf{W}^T ({\overline{\textbf{X}}}_i-{\overline{\textbf{X}}})\ |_F^2}{\sum\limits_{i=1}^{c}\sum\limits_{s=1}^{N_i}\|\textbf{W}^T(\textbf{X}_{is}-{\overline{\textbf{X}}_i})\|_F^2},
\end{split}
\end{equation}
where 
$\textbf{W}=(\textbf{w}_1,\ldots,\textbf{w}_r)\in\mathbb{R}^{d_1\times r}$, $r\leq d_1$.
$i=1,\ldots,c$, $j=1,\ldots,N_i$.
\eqref{2DLDA} can be solved through the generalized eigenvalue problem $\textbf{S}_b\textbf{w}=\lambda \textbf{S}_w\textbf{w}$ in case $\textbf{S}_w$ is nonsingular, and its solution is the $r$ eigenvectors corresponding to the first largest $r$ nonzero eigenvalues. After obtaining optimal $\textbf{W}$, a new sample $\textbf{X}\in\mathbb{R}^{d_1\times d_2}$ is projected into $\mathbb{R}^{r\times d_2}$ by $\textbf{W}^T\textbf{X}$.
Note that 2DLDA still will encounter the singularity problem when $\textbf{S}_w$ is not of full rank.

\section{Two-dimensional Bhattacharyya bound linear discriminant analysis}\label{sec2DBLDA}
In this section, we derive a new two-dimensional linear discriminant analysis criterion through minimizing a Bhattacharyya error bound.

\subsection{The derivation of a Bhattacharyya error bound estimation}

As we know, from the viewpoint of minimizing the probability of classification error, the Bayes classifier is the best classifier \cite{Fukunaga90}, and its error rate, or known as the Bayes error, is defined as
\begin{equation}\label{bayesoo}
\epsilon = 1- \int\underset{i\in\{1,2,\ldots,c\}}{max}\{P_ip_i(\textbf{X})\}d\textbf{X},
\end{equation}
where $\textbf{X}$ is a sample, $P_i$ and $p_i(\textbf{X})$ are the prior probability and the probability density function of the $i$-th class of the data, respectively.
The computation of the Bayes error is very difficult in general, and an alternative way of minimizing the Bayes error is to minimize its upper bound \cite{Saon02,RuedaHerrera08,Nielsen14}.
Bhattacharyya error \cite{Bhattacharyya43} provides a close upper bound to the Bayes error, which is given by
\begin{equation}\label{Berror}
\epsilon_B=\sum\limits_{i<j}^c\sqrt{P_iP_{j}}\int\sqrt{p_i(\textbf{X})p_{j}(\textbf{X})}d\textbf{X}.\\
\end{equation}

Under the background of two-dimensional supervised dimensionality reduction, if we can derive a relatively close upper bound of $\epsilon_B$, we may obtain a reasonable dimensionality reduction model. In fact, under some basic assumptions, we can obtain an upper bound of $\epsilon_B$, as shown in the following proposition.

\begin{prop}\label{propL2}
Assume $P_i$ and $p_i(\textbf{X})$ are the prior probability and the probability density function of the $i$-th class for the training data set $T$, respectively, and the data samples in each class are independent and identically normally distributed. Let $p_1(\textbf{X}), p_2(\textbf{X}),\ldots, p_c(\textbf{X})$ be the Gaussian functions given by $p_i(\textbf{X})=\mathcal{N}(\textbf{X}|{\overline{\textbf{X}}}_i, \boldsymbol{\Sigma}_i)$, where ${\overline{\textbf{X}}}_i$ and $\boldsymbol{\Sigma}_i$ are the class mean and the class covariance matrix, respectively. We further suppose $\boldsymbol{\Sigma}_i=\boldsymbol{\Sigma}$, $i=1,2,\ldots,c$, where $\boldsymbol{\Sigma}$ is the covariance matrix of the data set $T$, and ${\overline{\textbf{X}}}_i$ and $\boldsymbol{\Sigma}$ can be estimated accurately from $T$. Then for arbitrary projection vector $\textbf{w}\in\mathbb{R}^{d_1}$,
the Bhattacharyya error bound $\epsilon_B$ defined by \eqref{Berror} on the data set $\widetilde{T}=\{\widetilde{\textbf{X}}_i|\widetilde{\textbf{X}}_i=\textbf{w}^T\textbf{X}_i\in\mathbb{R}^{1\times d_2}\}$ satisfies the following:
\begin{equation}\label{BhattacharyyaL2}
\begin{split}
\epsilon_B
\leq&-\frac{a}{8}\sum\limits_{i<j}^c\sqrt{P_iP_j}{||\textbf{w}^T({\overline{\textbf{X}}_{i}-\overline{\textbf{X}}_{j}})||_2^2}+\frac{a}{8}\Delta\sum_{i=1}^{c}\sum_{s=1}^{N_i}||\textbf{w}^T(\textbf{X}_{is}-\overline{\textbf{X}}_i)||_2^2\\
&+\sum\limits_{i<j}^c\sqrt{P_iP_j},\\
\end{split}
\end{equation}
where $\Delta=\frac{1}{4}\sum\limits_{i<j}^c\sqrt{P_iP_j}||{\overline{\textbf{X}}_{i}-\overline{\textbf{X}}_{j}}||_F^2$, and $a>0$ is some constant.
\end{prop}
\noindent \textbf{Proof:}  See the Appendix. \hfill $\square$

\subsection{The proposed two-dimensional Bhattacharyya bound linear discriminant analysis}

Proposition \ref{propL2} gives a reasonable upper bound of $\epsilon_B$.
After obtaining an upper error bound, it is natural to minimize it. Therefore, we minimize the upper bound of $\epsilon_B$ in \eqref{BhattacharyyaL2}, that is, the right side of \eqref{BhattacharyyaL2}. In fact, by minimizing it, we can easily obtain a novel two-dimensional Bhattacharyya bound linear discriminant analysis (2DBLDA) as the following
\begin{equation}
\begin{split}\label{2DBLDAw}
\underset{\textbf{w}^T\textbf{w}=1}{\min}~~&-\frac{1}{N}\sum_{i<j}\sqrt{N_iN_j}||\textbf{w}^T(\overline{\textbf{X}}_{i}-\overline{\textbf{X}}_{j})||_2^2+\Delta\sum_{i=1}^{c}\sum_{s=1}^{N_i}||\textbf{w}^T(\textbf{X}_{is}-\overline{\textbf{X}}_i)||_2^2\\
\end{split}
\end{equation}
where $\Delta=\frac{1}{4}\sum\limits_{i<j}^c\sqrt{P_iP_j}||{\overline{\textbf{X}}_{i}-\overline{\textbf{X}}_{j}}||_F^2$,
$\textbf{w}\in\mathbb{R}^{d_1}$, $P_i=\frac{N_i}{N}$.

By applying \eqref{2DBLDAw}, we can project a $d_1\times d_2$ sample $\textbf{X}$ into a $1\times d_2$ sample $\widetilde{\textbf{X}}$ by $\widetilde{\textbf{X}}=\textbf{w}^T\textbf{X}$. However, it not usually contains enough discriminant information in the $1\times d_2$ space, and we may need $r\geq 1$ projection vectors $\textbf{w}_1, \textbf{w}_2,\ldots,\textbf{w}_r$ that constitute a projection matrix $\textbf{W}=(\textbf{w}_1, \textbf{w}_2,\ldots,\textbf{w}_r)\in\mathbb{R}^{d_1\times r}$, and project $\textbf{X}$ into a $r\times d_2$ space by $\widetilde{\textbf{X}}=\textbf{W}^T\textbf{X}$.

In general, we consider the following 2DBLDA
\begin{equation}
\begin{split}\label{2DBLDA}
\underset{\textbf{W}}{\min}~~&-\frac{1}{N}\sum_{i<j}\sqrt{N_iN_j}||\textbf{W}^T(\overline{\textbf{X}}_{i}-\overline{\textbf{X}}_{j})||_F^2+\Delta\sum_{i=1}^{c}\sum_{s=1}^{N_i}||\textbf{W}^T(\textbf{X}_{is}-\overline{\textbf{X}}_i)||_F^2\\
\hbox{s.t.\ }& \textbf{W}^T\textbf{W}=\textbf{\textbf{I}},
\end{split}
\end{equation}
where $\textbf{W}\in\mathbb{R}^{r\times d_1}$, $r\leq d_1$.
We now give the geometric meaning of 2DBLDA.
Minimizing the first term in \eqref{2DBLDA} will make the means of two different classes far from each other in the projected space, which guarantees the between-class separativeness. Here the coefficients $\frac{1}{N}\sqrt{N_iN_j}$ in the first term weight distance pairs between different class means.
Minimizing the second term in \eqref{2DBLDA} forces each sample around its own class mean in the projected space.
The weighting constant $\Delta$ in front of the second term balances the between-class importance and within-class importance while also makes sure minimum error bound according to the proof of Proposition 1. We can observe that 2DBLDA is adaptive to different data since $\Delta$ is determined by the given data set.
The constraint $\textbf{W}^T\textbf{W}=\textbf{I}$ makes sure the obtained discriminant directions orthogonormal to each other, which ensures minimum redundancy in the projected space.

2DBLDA can be easily solved through a simple standard eigenvalue decomposition problem. In fact, model \eqref{2DBLDA} can be rewritten as
\begin{equation}
\begin{split}\label{L2BLDA2}
\underset{\textbf{W}}{\min}&~~tr(\textbf{W}^T\textbf{S}\textbf{W})\\
\hbox{s.t.\ }& \textbf{W}^T\textbf{W}=\textbf{I},
\end{split}
\end{equation}
where
\begin{equation}
\begin{split}
\textbf{S}=&-\frac{1}{N}\sum_{i<j}\sqrt{N_iN_j}(\overline{\textbf{X}}_{i}-\overline{\textbf{X}}_{j})(\overline{\textbf{X}}_{i}-\overline{\textbf{X}}_{j})^T+\Delta\sum_{i=1}^{c}\sum_{s=1}^{N_i}(\textbf{X}_{is}-\overline{\textbf{X}}_i)(\textbf{X}_{is}-\overline{\textbf{X}}_i)^T.
\end{split}
\end{equation}
Then $\textbf{W}=\left(\textbf{w}_1,\textbf{w}_2,\ldots,\textbf{w}_r\right)$ is obtained by the $r$ orthogonormal eigenvectors that correspond to the first $r$ smallest nonzero eigenvectors of $\textbf{S}$.

\section{Experiments}\label{secExp}
In this section, we compare our 2DBLDA with 2DPCA \cite{YangZhang04}, 2DPCA-L1 \cite{LiPangYuan10}, 2DLDA \cite{XiongSwamy05} and L1-2DLDA \cite{L12DLDA15,ChenL12DLDA15}. The learning parameter $\delta$ of L1-2DLDA is selected optimally from the set $\{0.001, 0.005, 0.01, 0.05, 0.1, 0.5, 1\}$ by grid search.
We experiment on Yale database\footnote{http://cvc.cs.yale.edu/cvc/projects/yalefaces/yalefaces.html} and Coil100 database \cite{NeneNayar96} for image recognition, and on Indian females database \footnote{http://www.cs.umass.edu/{\~{}}vidit/IndianFaceDatabase} for face reconstruction.
After applying the above dimensionality reduction methods on training data, the test data are projected to lower dimensional space. For image recognition, the nearest neighbors classifier is employed to obtain classification accuracy as the performance measurement. For face reconstruction, the mean reconstruction error is used for performance evaluation. All the methods are carried out on a PC with P4 2.3 GHz CPU by Matlab 2017b.
\subsection{Image recognition}
In this subsection, we apply the proposed method on the Yale and Coil100 databases for image recognition.
The Yale database is a human face database that contains 165 images of 15 individuals, and each individual includes 11 images. The database is considered to evaluate the performance of methods when facial expression and lighting conditions are changed. The Coil100 database contains 900 images of 100 objects, with each object containing 9 images. For Yale database, 7 images for each individual are randomly chosen to form the training set and the rest images compose the test set. For Coil100 database, 6 images for each subject are randomly chosen to form the training set and the rest images construct the test set.

We first apply all the methods on the original training data and obtain their projection matrices. The test classification accuracies on the obtained projected test data of these two databases are listed in Table \ref{TableOri}, and the best accuracies are shown in bold figure. For 2DPCA-L1 and L1-2DLDA, since their performance is affected by the initialization projections, we repeat these two methods ten times and adopt their mean accuracies along with standard variances. From the table, we see our 2DBLDA owns the highest or comparable performance compared to other methods.
\begin{table*}[htbp]
\begin{center}
\caption{Comparison of mean accuracy (\%) for different methods on the original Yale database and Coil100 database.}
\resizebox{4.5in}{!}
{
\begin{tabular}{l|ccccccccccc}
\toprule
{\multirow{1}{*}{Data set~}} & 2DPCA~~ & 2DPCA-L1~~ & 2DLDA~~ & L1-2DLDA~~ & 2DBLDA \\
\midrule
Yale~~ &\textbf{85.00} &82.67$\pm$0.86 &83.33  &\textbf{85.00}$\pm$0.00 &  \textbf{85.00}\\
Coil100~~ &74.00  &67.93$\pm$1.26  &72.00 &73.37$\pm$0.46  &  \textbf{74.33} \\
\bottomrule
\end{tabular}
}
\label{TableOri}
\end{center}
\end{table*}

\begin{figure}[htbp]
\begin{center}{
\subfigure[Yale samples]{
\resizebox*{6.5cm}{!}
{\includegraphics{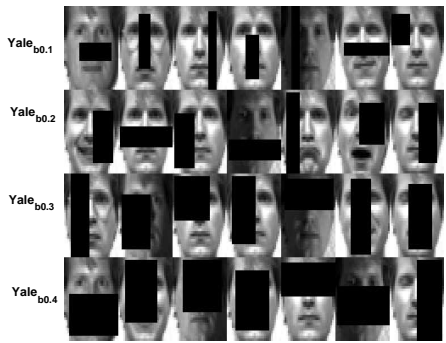}}}\hspace{5pt}
\subfigure[Coil100 samples]{
\resizebox*{6.5cm}{!}
{\includegraphics{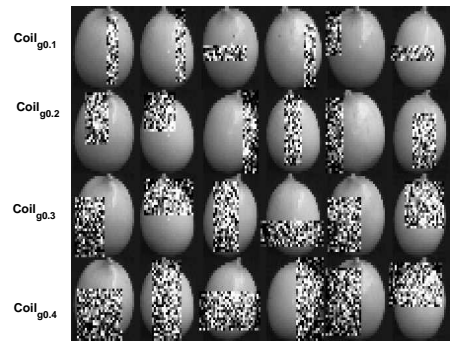}}}\hspace{5pt}
\caption{Noise samples from the Yale and Coil100 databases.}
\label{FigNoiseImage}}
\end{center}
\end{figure}

To further see the superiority of our 2DBLDA, we artificially pollute the training data by adding each training sample with a rectangle block occlusion at a random location. We here set the occlusion area ratio to ${10\%, 20\%, 30\%, 40\%}$, respectively. For convenience, we denote these four data sets as Yale$_{b0.1}$, Yale$_{b0.2}$, Yale$_{b0.3}$ and Yale$_{b0.4}$, where the subscript ``$b$" represents block occlusion and the number next to it means occlusion ratio. For the Coil100 data, we add random rectangular Gaussian noise of mean 0 and variance 0.2 that covers ${10\%, 20\%, 30\%, 40\%}$ areas of each training image at random position. We denote these four data sets as Coil$_{g0.1}$, Coil$_{g0.2}$, Coil$_{g0.3}$ and Coil$_{g0.4}$, where the subscript ``$g$" represents Gaussian noise and the number next to it means noise ratio. Some noise samples are shown in Fig.\ref{FigNoiseImage}.
The classification results on the noise data sets are listed in Tables \ref{TableYaleNoise} and \ref{TableCoilNoise} respectively. From the tables, we have the following observations:
(i) All methods are affected by the noise, and their corresponding accuracies are lower than the ones on the original data. In general, the larger the noise area is, the lower the accuracy is. (ii) The proposed 2DBLDA owns the highest accuracy on all noise data. (iii) L1-2DLDA and 2DPCA perform better than 2DPCA-L1 and 2DLDA.

\begin{table*}[htbp]
\begin{center}
\caption{Comparison of mean accuracy (\%) for different methods on noise Yale databases.}

\resizebox{4.5in}{!}
{
\begin{tabular}{l|ccccccccccc}
\toprule
{\multirow{1}{*}{Data set}} & 2DPCA~~ & 2DPCA-L1~~ & 2DLDA~~ & L1-2DLDA~~ & 2DBLDA \\
\midrule
Yale$_{b0.1}$ &76.67 &59.33$\pm$2.63  &76.66   &77.33$\pm$1.17  &  \textbf{78.33}  \\
Yale$_{b0.2}$  &\textbf{76.67}   & 55.83$\pm$3.07 & 70.00  &73.67$\pm$2.33 & \textbf{76.67}  \\
Yale$_{b0.3}$  &63.33   &50.67$\pm$7.58  &63.33   &64.67$\pm$2.33 & \textbf{65.00}  \\
Yale$_{b0.4}$  &56.67   &49.67$\pm$3.31 &56.67   &51.83$\pm$3.46  &\textbf{60.00}  \\
\bottomrule
\end{tabular}
}
\label{TableYaleNoise}
\end{center}
\end{table*}

\begin{table*}[htbp]
\begin{center}
\caption{Comparison of mean accuracy (\%) for different methods on  noise Coil100 database.}

\resizebox{4.5in}{!}
{
\begin{tabular}{l|ccccccccccc}
\toprule
{\multirow{1}{*}{Data set}} & 2DPCA~~ & 2DPCA-L1~~ & 2DLDA~~ & L1-2DLDA~~ & 2DBLDA \\
\midrule
Coil$_{g0.1}$ &71.67   &63.77$\pm$3.39  &70.67   &71.37$\pm$1.11  &  \textbf{72.33}  \\
Coil$_{g0.2}$  &67.00   &51.43$\pm$1.75 &68.67   &67.10$\pm$0.88 &\textbf{70.00}   \\
Coil$_{g0.3}$  &63.33  &48.93$\pm$3.43  &63.33    &61.80$\pm$2.54 &\textbf{68.33}   \\
Coil$_{g0.4}$  &58.00   & 43.70$\pm$0.64 &60.00   &57.73$\pm$1.84  &\textbf{61.00}  \\
\bottomrule
\end{tabular}
}
\label{TableCoilNoise}
\end{center}
\end{table*}

\begin{figure}[htbp]
\begin{center}{
\subfigure[Yale]{
\resizebox*{6.5cm}{!}
{\includegraphics{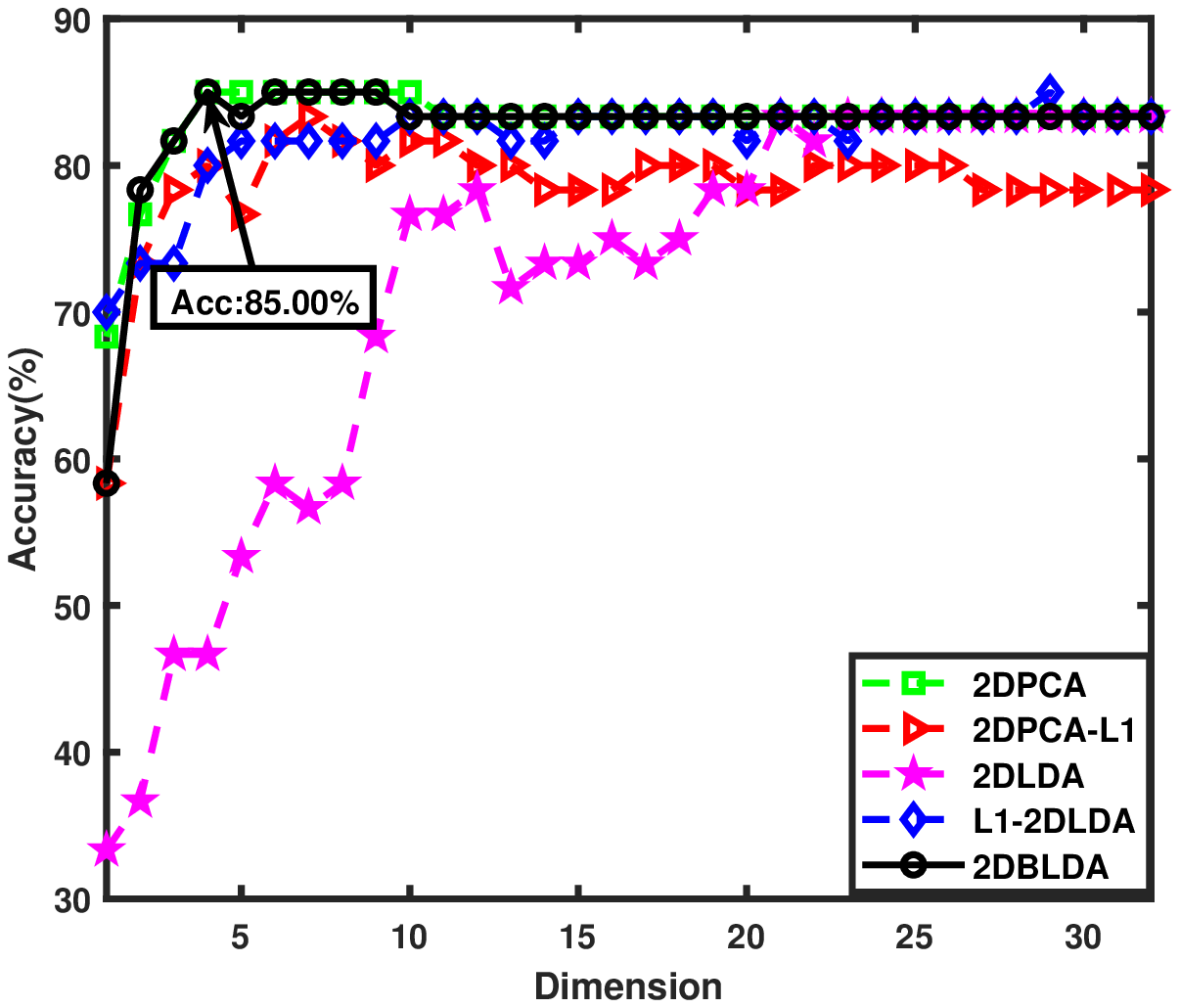}}}\hspace{5pt}
\subfigure[Coil100]{
\resizebox*{6.5cm}{!}
{\includegraphics{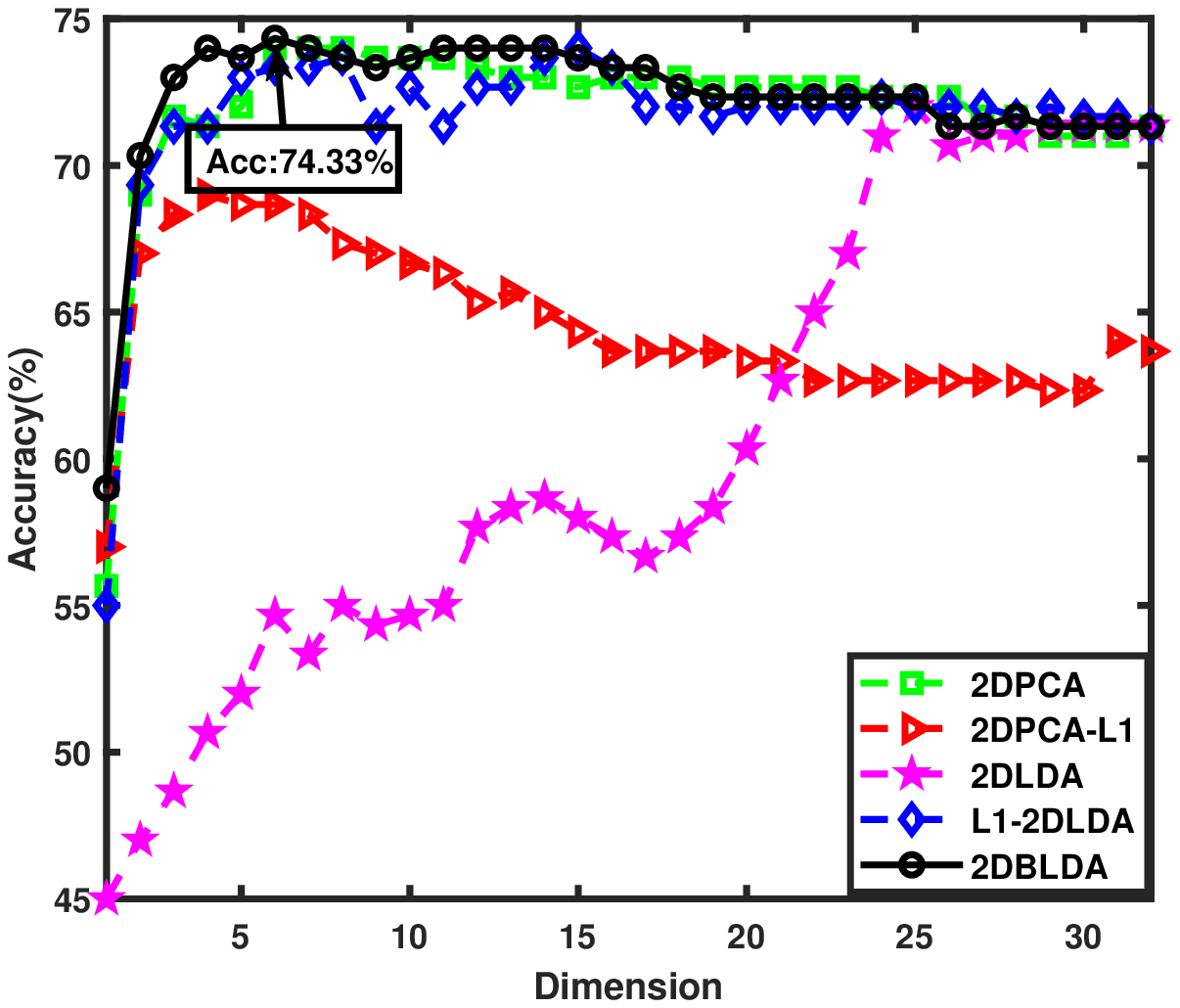}}}\hspace{5pt}
\caption{Accuracies of all methods on original Yale and Coil100 databases.}
\label{FigDimV0}}
\end{center}
\end{figure}

\begin{figure}[htbp]
\begin{center}{
\subfigure[Yale$_{b0.1}$]{
\resizebox*{6.5cm}{!}
{\includegraphics{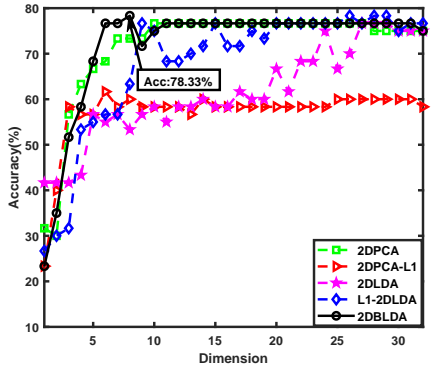}}}\hspace{5pt}
\subfigure[Yale$_{b0.2}$]{
\resizebox*{6.5cm}{!}
{\includegraphics{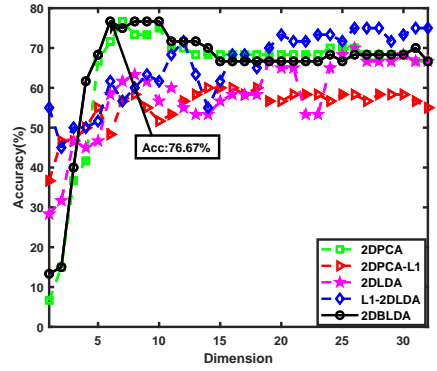}}}\hspace{5pt}
\subfigure[Yale$_{b0.3}$]{
\resizebox*{6.5cm}{!}
{\includegraphics{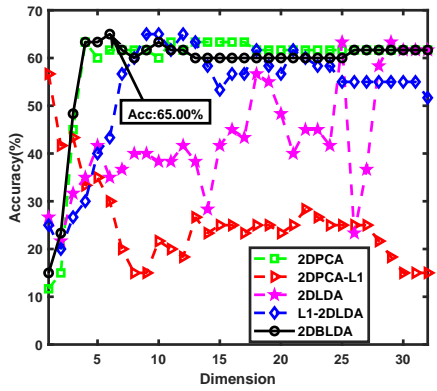}}}\hspace{5pt}
\subfigure[Yale$_{b0.4}$]{
\resizebox*{6.5cm}{!}
{\includegraphics{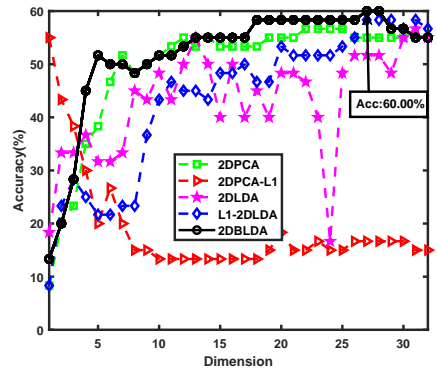}}}\hspace{5pt}
\caption{Accuracies of all methods on noise Yale database.}
\label{FigDimV1}}
\end{center}
\end{figure}

\begin{figure}[htbp]
\begin{center}{
\subfigure[Coil$_{g0.1}$]{
\resizebox*{6.5cm}{!}
{\includegraphics{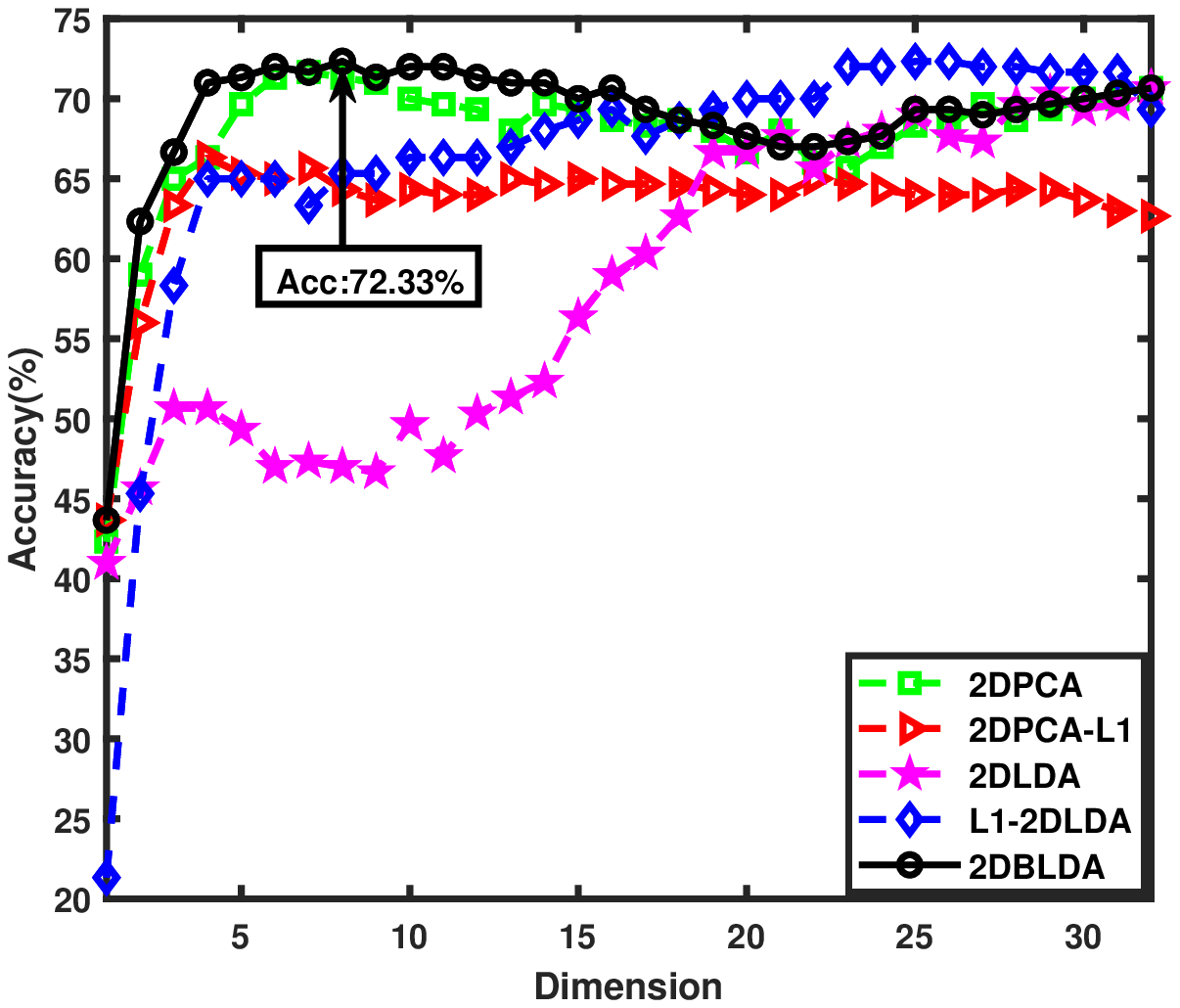}}}\hspace{5pt}
\subfigure[Coil$_{g0.2}$]{
\resizebox*{6.5cm}{!}
{\includegraphics{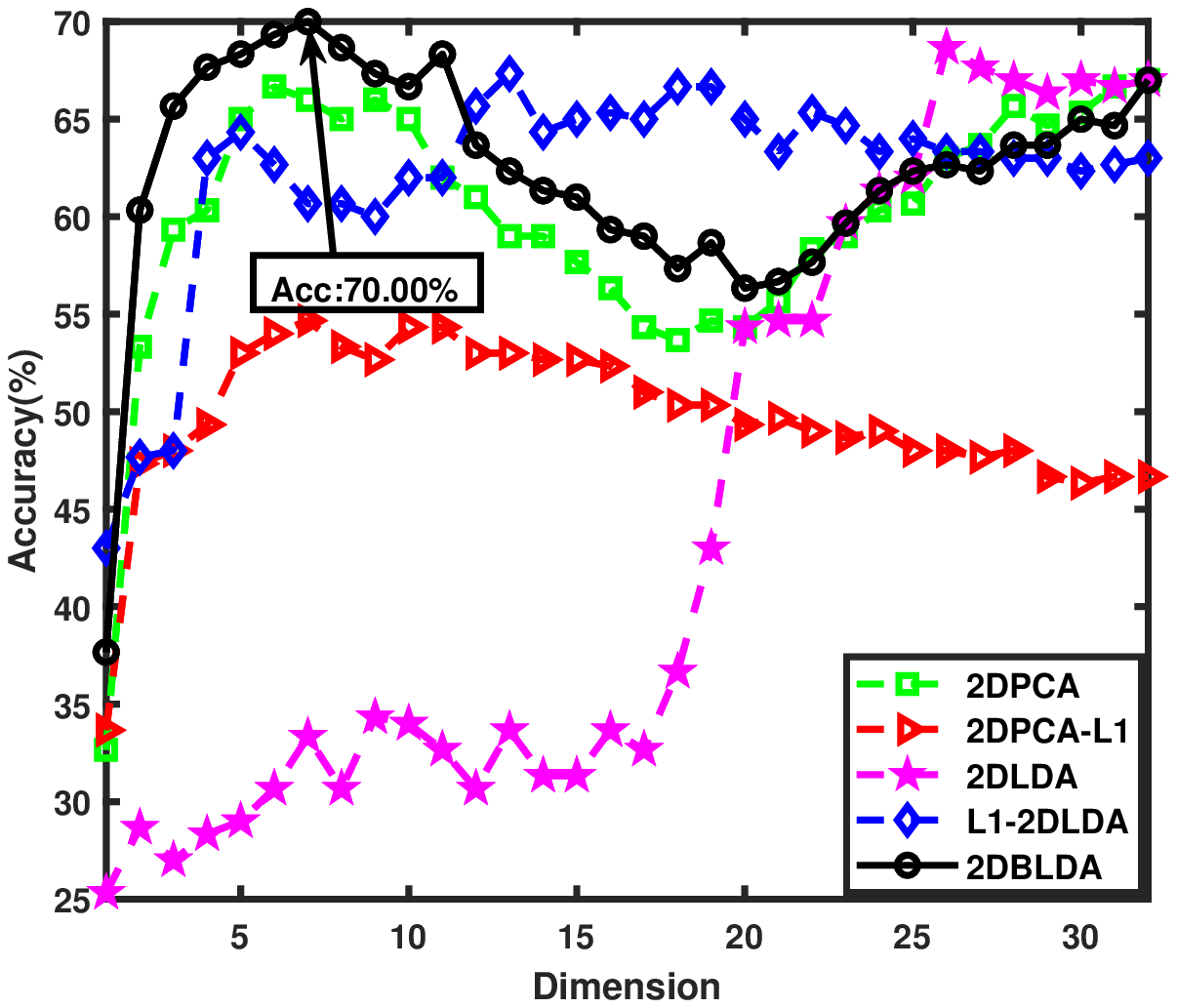}}}\hspace{5pt}
\subfigure[Coil$_{g0.3}$]{
\resizebox*{6.5cm}{!}
{\includegraphics{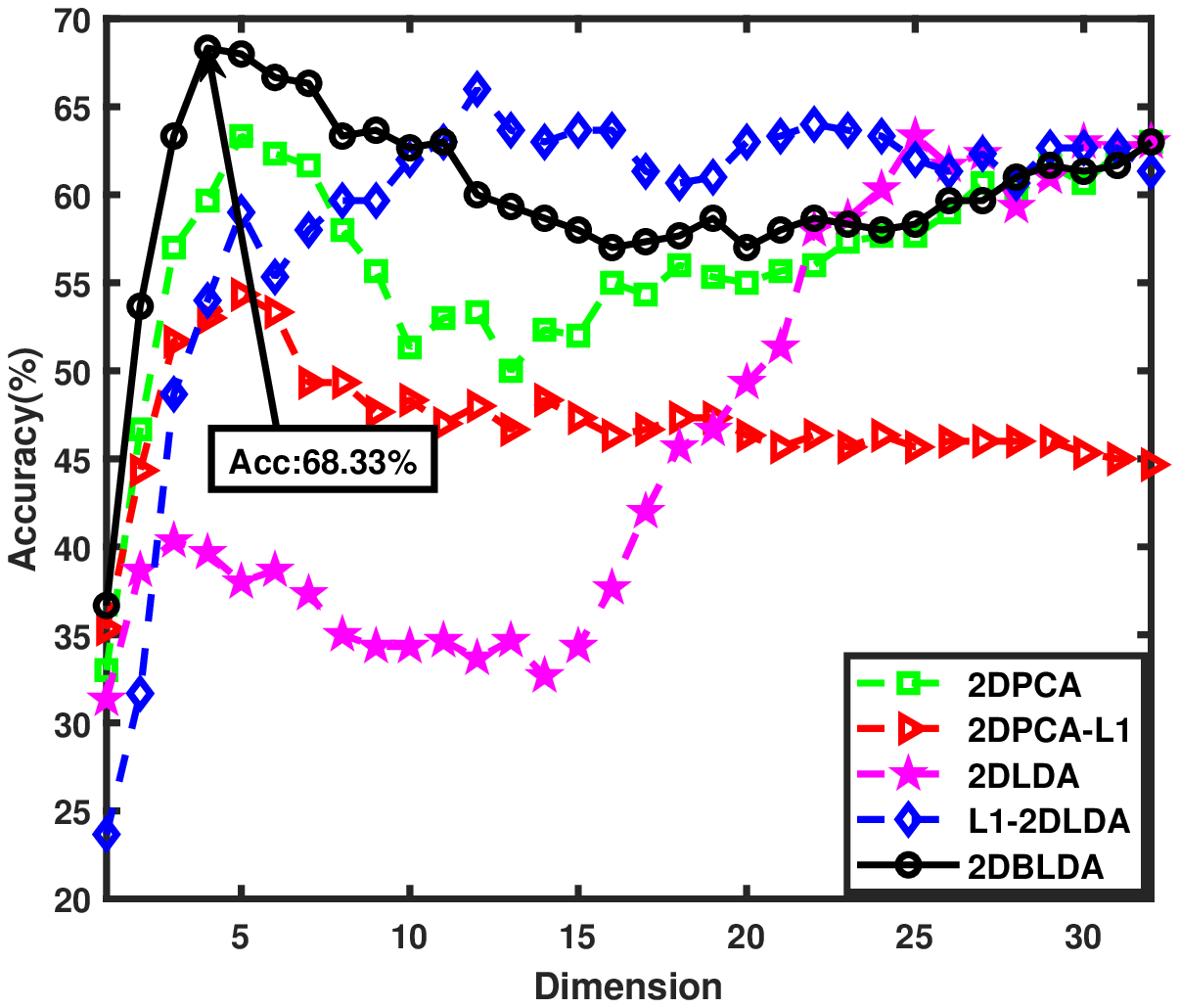}}}\hspace{5pt}
\subfigure[Coil$_{g0.4}$]{
\resizebox*{6.5cm}{!}
{\includegraphics{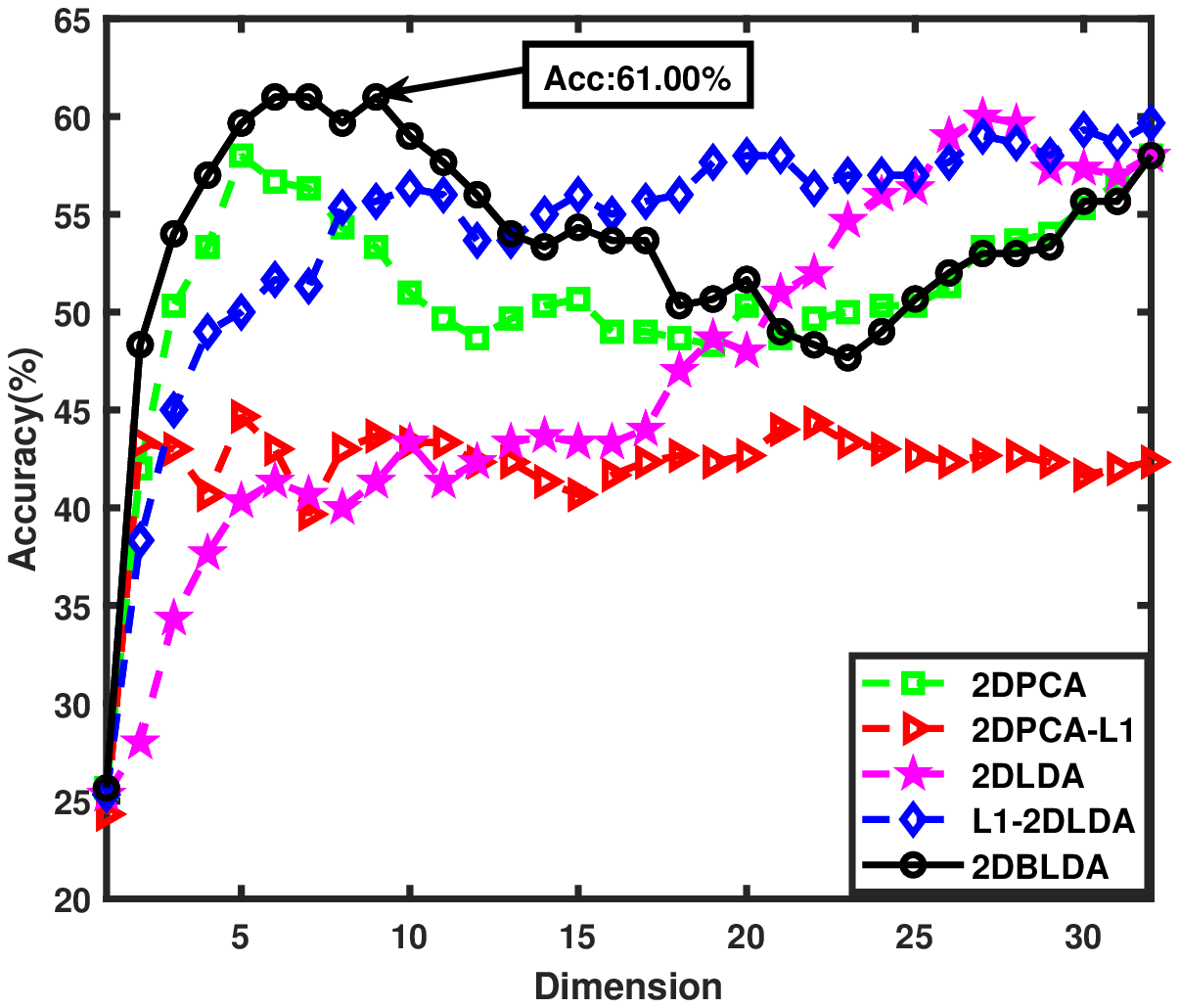}}}\hspace{5pt}
\caption{ Accuracies of all methods on noise Coil100 database.}
\label{FigDimV2}}
\end{center}
\end{figure}

We further investigate the influence of the reduced dimension to the accraucy. Fig.\ref{FigDimV0} depicts the variation of accuracies along dimensions on original Yale and Coil100 databases, and Figs.\ref{FigDimV1} and \ref{FigDimV2} depict the corresponding results on noise databases.
The results show that: (i) As the increasing of the number of reduced dimensions, the accuracies of 2DPCA and our 2DBLDA first achieve their highest and then have a relative steady trend, while other methods vary a lot. (ii) No matter on the original data or the noise data, the proposed 2DBLDA has the highest accuracy under optimal reduced dimension. (iii) All the methods are greatly influenced by the reduced dimension, and it is necessary to choose an optimal reduced dimension. (iv) In addition, the optimal reduced dimension of 2DBLDA is not too large compared to other methods in general.

\subsection{Face Reconstruction}
In this part, we apply the proposed 2DBLDA and other methods to face reconstruction on the Indian females database. The Indian females database contains 242 human face images of 22 female individuals and each individual has 11 different images. The original images are resized to 32$\times$32 pixels.
We first introduce face image reconstruction.
For a given image $\textbf{X}\in\mathbb{R}^{d_1\times d_2}$, suppose we have obtained a projection matrix $\textbf{W}=\left(\textbf{w}_1,\textbf{w}_2,\ldots,\textbf{w}_r\right)\in\mathbb{R}^{d_1\times r}$, $r\leq d_1$. Then $\textbf{X}$ is projected into the ${r\times d_2}$-dimensional space by $\widetilde{\textbf{X}}=\textbf{W}^T\textbf{X}$. Since $\textbf{w}_1,\textbf{w}_2,\ldots,\textbf{w}_r$ are orthonormal, the reconstructed image of $\textbf{X}$ can be obtained by $\widehat{\textbf{X}}=\textbf{W}\widetilde{\textbf{X}}=\textbf{W}\textbf{W}^T\textbf{X}$.
To measure the reconstruction performance, we use the average reconstruction error (ARE) as a measure, which is defined as
\begin{equation}\label{error}
\bar{e}_{r}=\frac{1}{N}\sum\limits_{i=1}^{N}||\textbf{X}_i-\textbf{W}\textbf{W}^T\textbf{X}_i||_{F},
\end{equation}
where $r=1,2,\ldots,d_1$.

\begin{figure}[htbp]
\begin{center}{
\subfigure[Average reconstruction error]{
\resizebox*{6.6cm}{!}
{\includegraphics{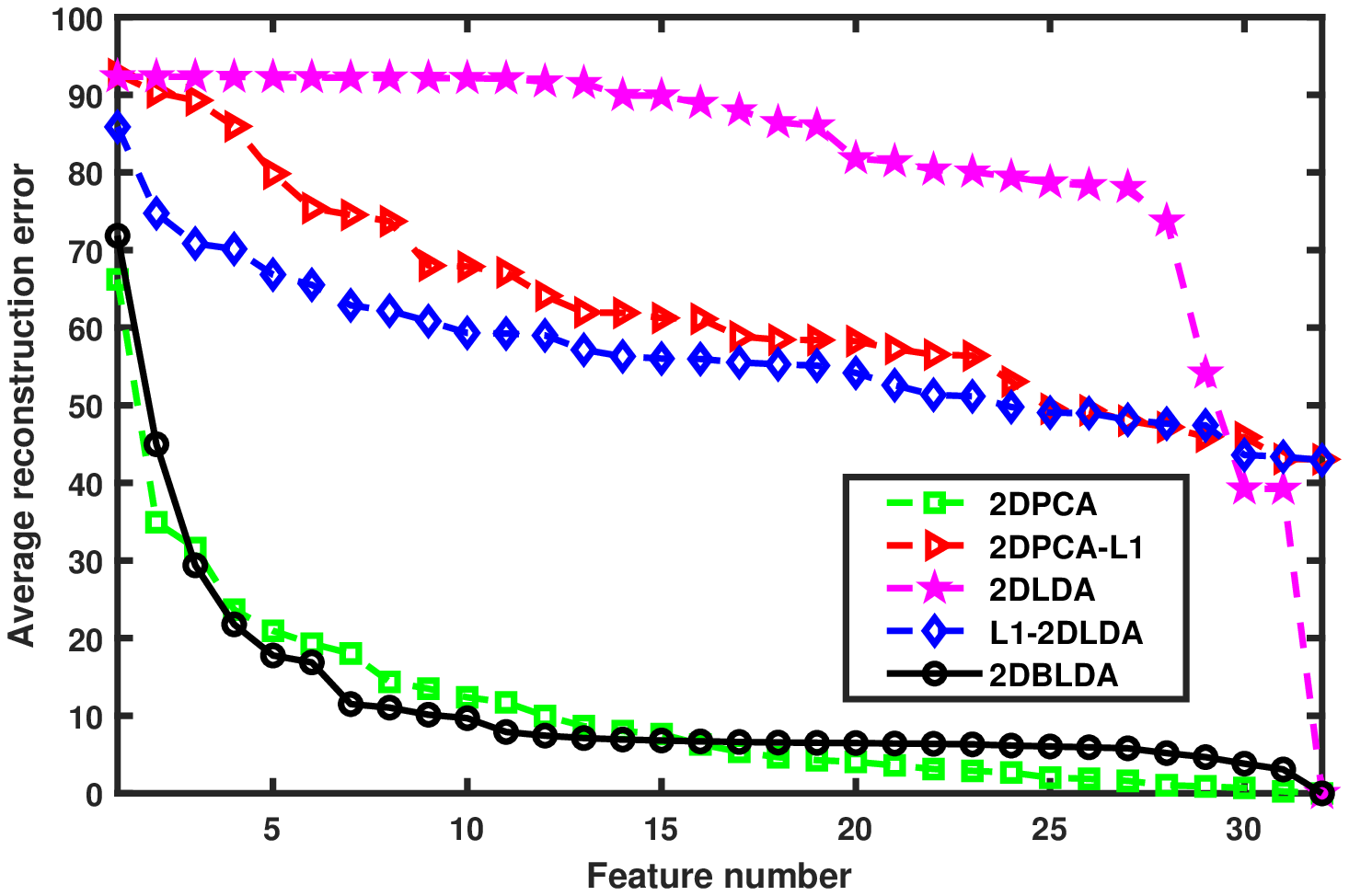}}}\hspace{5pt}
\subfigure[Reconstructed faces]{
\resizebox*{6.5cm}{!}
{\includegraphics{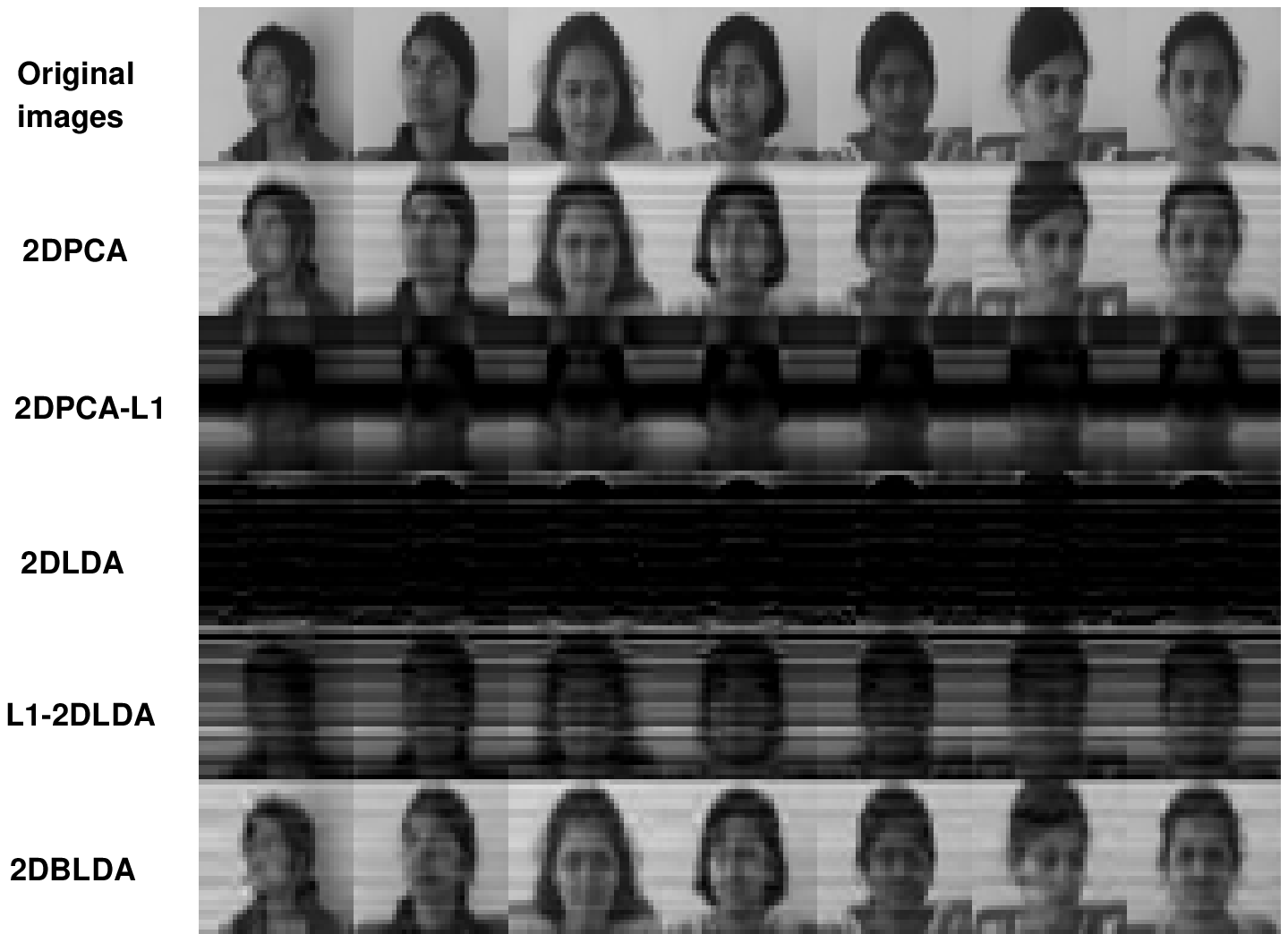}}}\hspace{5pt}
\caption{Face reconstruction results on original Indian database.}
\label{FigRecOrig}}
\end{center}
\end{figure}

We first experiment on the original data, and compute the ARE for each method. The variation of ARE along different dimensions is shown in Fig.\ref{FigRecOrig} (a).
From the figure, we see when the dimension is less than 15, our 2DBLDA performs the best especially when the dimension is greater than 5. When the dimension is greater than 15, 2DPCA is comparable or slightly better than our 2DBLDA, but both of these two methods almost achieve steady. The result shows that 2DBLDA can achieve good performance for low dimensions.
Other three methods obviously perform worse than our 2DBLDA and 2DPCA on all the dimensions.
When $r=15$, we demonstrate the reconstructed face images for random 7 individuals in Fig.\ref{FigRecOrig} (b). It can be visually seen that 2DBLDA and 2DPCA have the best reconstruction effect.

\begin{figure}[htbp]
\begin{center}{
\subfigure[Average reconstruction error]{
\resizebox*{6.6cm}{!}
{\includegraphics{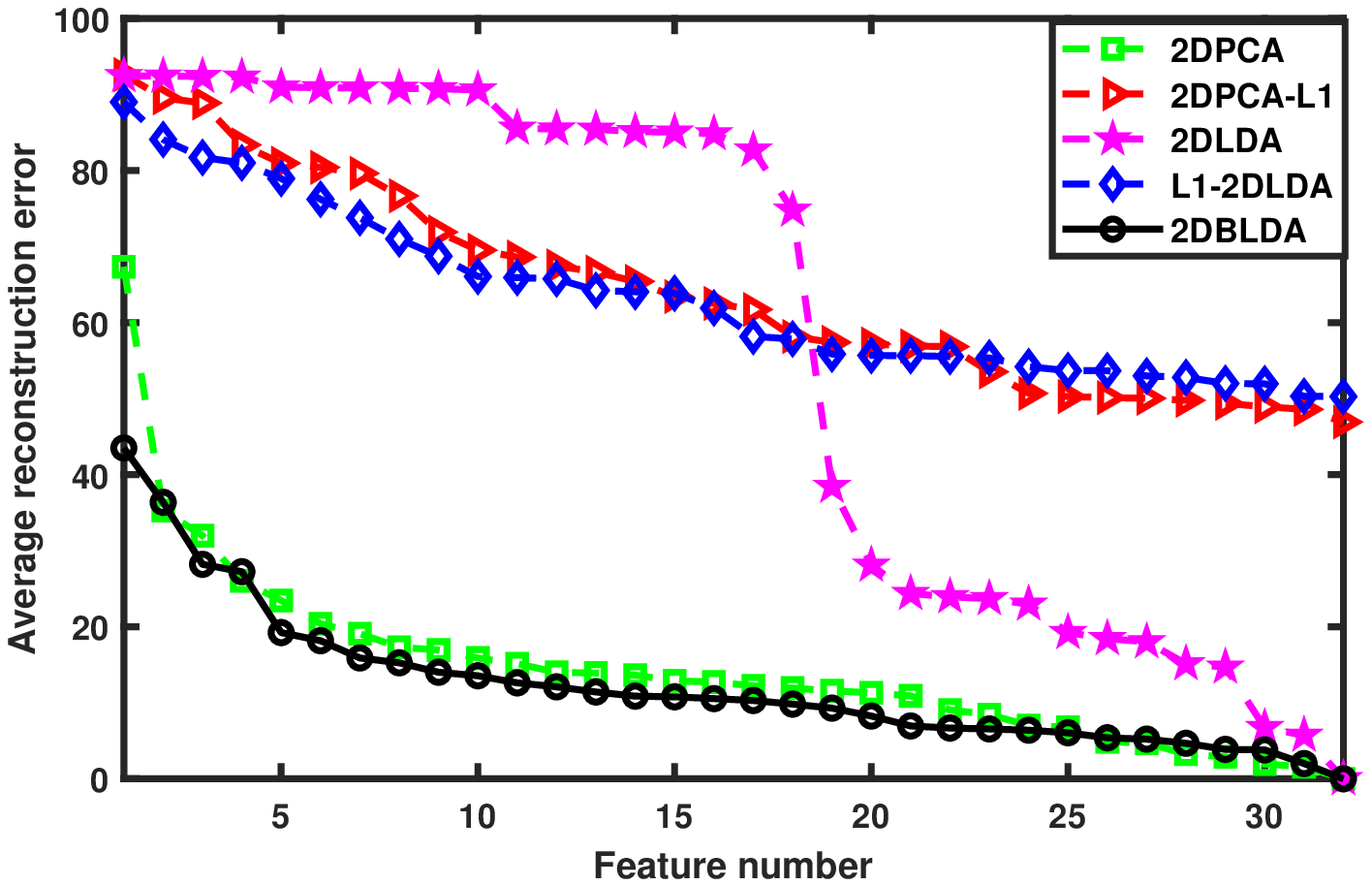}}}\hspace{5pt}
\subfigure[Reconstructed faces]{
\resizebox*{6.5cm}{!}
{\includegraphics{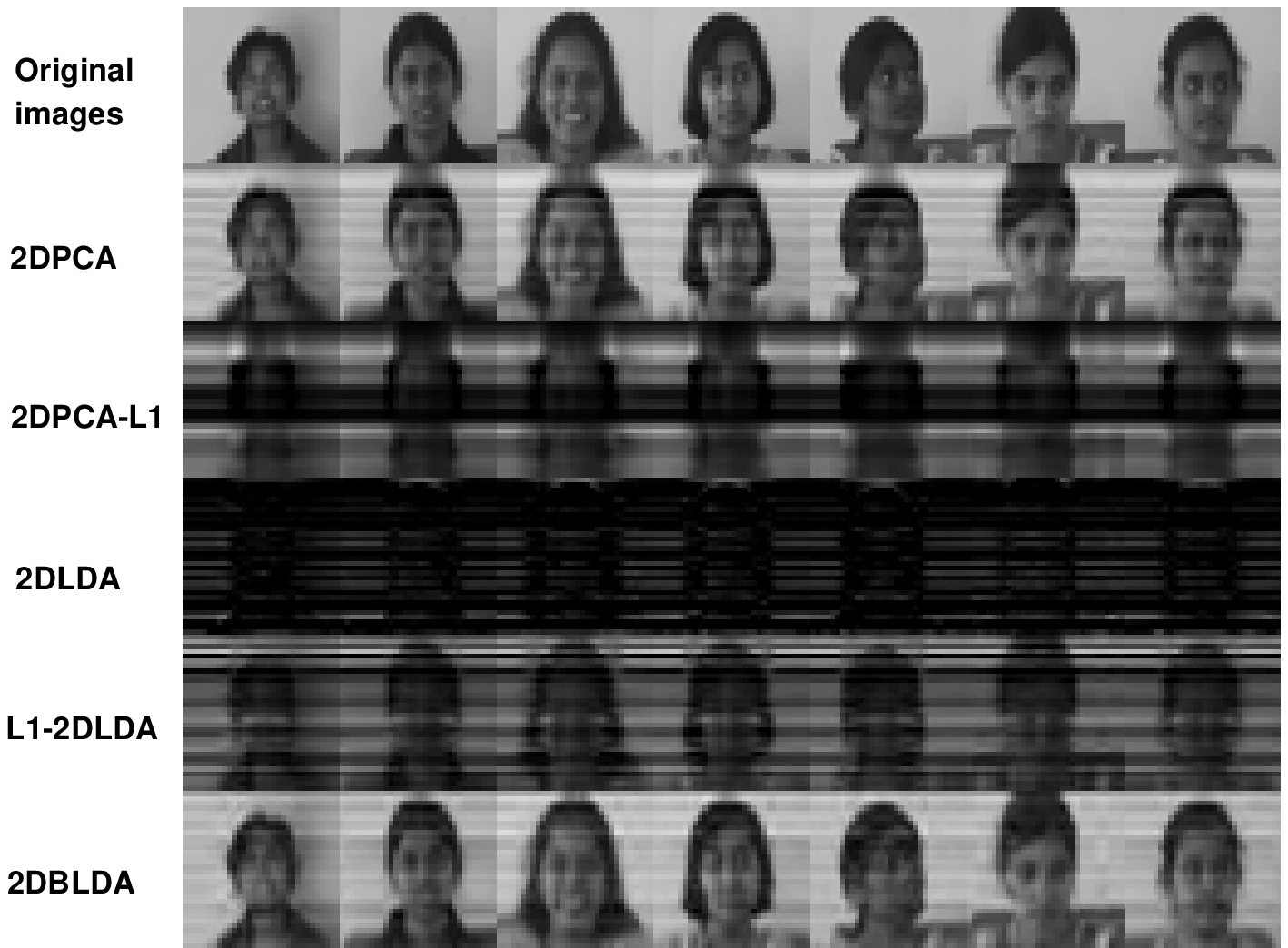}}}\hspace{5pt}
\caption{Face reconstruction results on Gaussian Indian database.}
\label{FigRecGau}}
\end{center}
\end{figure}

\begin{figure}[htbp]
\begin{center}{
\subfigure[Average reconstruction error]{
\resizebox*{6.6cm}{!}
{\includegraphics{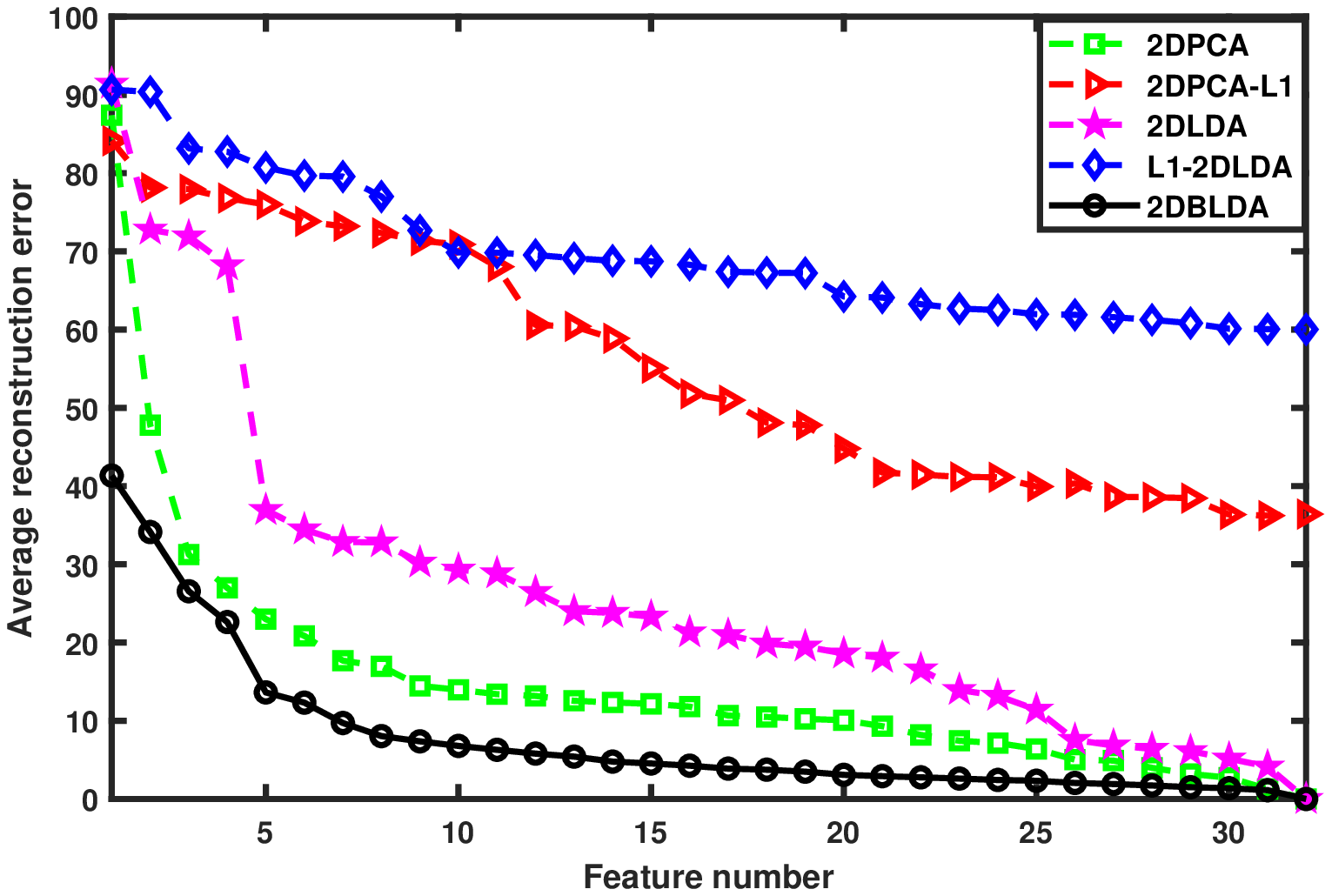}}}\hspace{5pt}
\subfigure[Reconstructed faces]{
\resizebox*{6.5cm}{!}
{\includegraphics{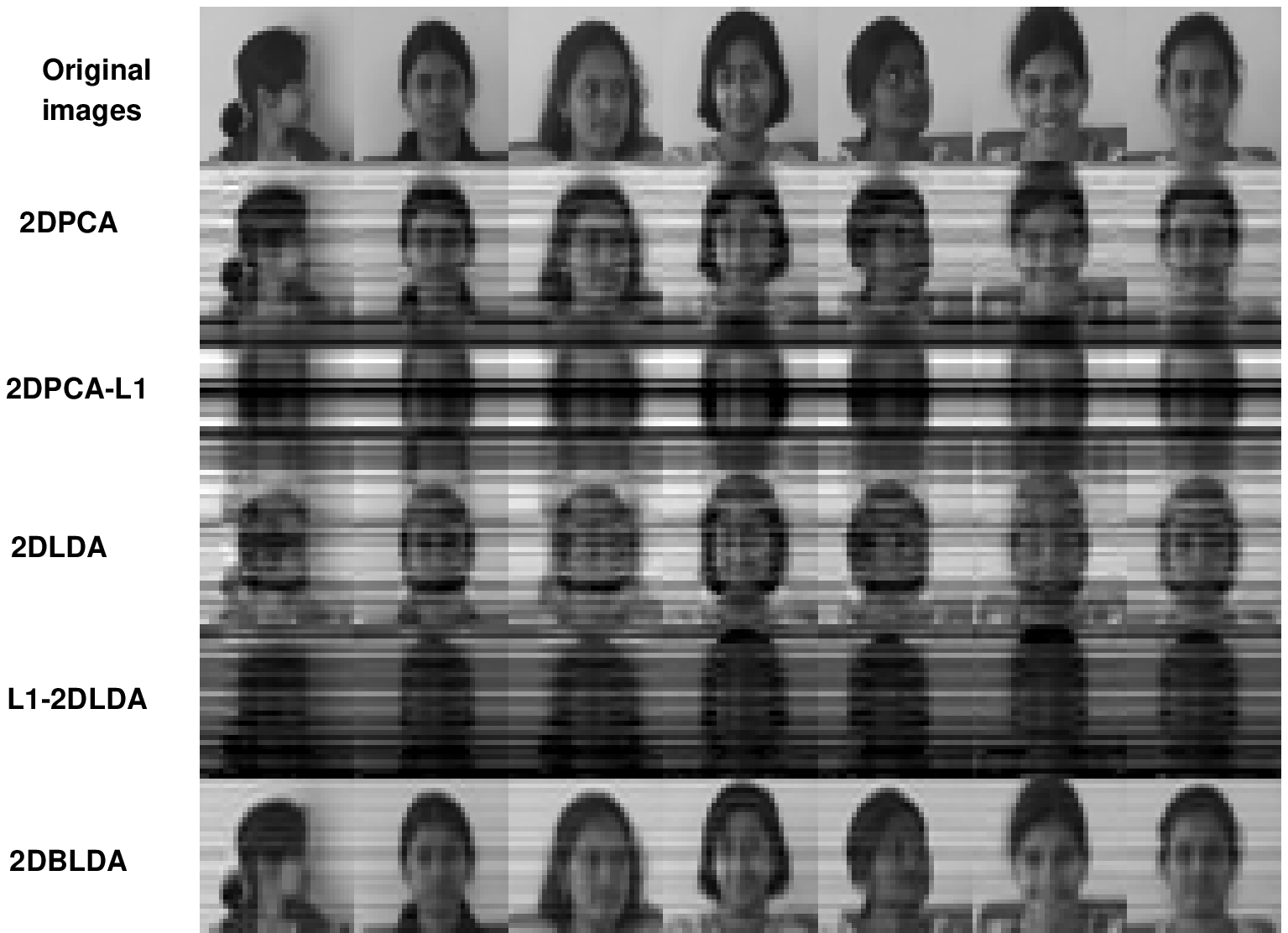}}}\hspace{5pt}
\caption{Face reconstruction results on dummy Indian database.}
\label{FigRecDum}}
\end{center}
\end{figure}

To further evaluate the effectiveness of the proposed 2DBLDA, we add two different types of noise on the data. The first type noise is the Gaussian noise of mean 0 and variance 0.05 that covers 30\% areas of each image.
The ARE of each method under different dimensions is plotted in Fig.\ref{FigRecGau}(a).
On this noise data, we see our 2DBLDA outperforms other methods on almost all the reduced dimensions, and 2DPCA is comparable to our 2DBLDA only when the dimension is greater than 27. This indicates that the proposed 2DBLDA can achieve fairly good performance by only employing a small number of reduced dimensions. We then add the second type of noise named dummy noise on the data. Here the dummy noise is the image which is generated from the discrete uniform distribution on [0,1], and is of the same size as the original image. Extra 100 dummy images are added to the whole database. After the projection matrix is obtained on this polluted data, it is used to reconstruct human face images. The result in Fig.\ref{FigRecDum} (a) demonstrates that our 2DBLDA has the lowest ARE on this database for all the dimensions, and when the dimension is greater than 20, it has rather low ARE. The reconstructed face images when $r=15$ that are shown in Fig.\ref{FigRecDum} (b) also support the above argument.

\section{Conclusion}\label{secCon}

This paper proposed a novel two-dimensional linear discriminant analysis via the Bhattacharyya upper bound optimality (2DBLDA). Different from the existing 2DLDA, optimizing the criterion of 2DBLDA was equivalent to optimizing an upper bound of the Bhattacharyya error. This led to maximizing a weighted between-class distance and minimizing the within-class distance, where these two distances were weighted by a meaningful adaptive constant that can be computed directly by the involved data. 2DBLDA had no parameters to be tuned and can be effectively solved by a standard eigenvalue decomposition problem. Experimental results on image recognition and face image reconstruction demonstrated the superiority of the proposed method. Our Matlab code can be downloaded from http://www.optimal-group.org/Resources/Code/2DBLDA.html.

\section*{Appendix}\label{AppendixBayes}
\noindent \textbf{Proof of Proposition 1:} We first note that $p_i(\widetilde{{\textbf{X}}})=\mathcal{N}(\widetilde{{\textbf{X}}}|\widetilde{{\overline{\textbf{X}}}}_i, \widetilde{\boldsymbol{\Sigma}})$, where
$\widetilde{{\overline {\textbf{X}}}}_i=\textbf{w}^T\overline{\textbf{X}}_i\in\mathbb{R}^{1\times d_2}$ is the $i$-class mean and $\widetilde{\boldsymbol{\Sigma}}$ is the covariance matrix in the $1\times d_2$ projected space.
Denote
\begin{equation}\label{D}
\textbf{D}=\begin{pmatrix}\textbf{w}^T\textbf{X}_1\\ \vdots\\\textbf{w}^T\textbf{X}_N\end{pmatrix}^T\in\mathbb{R}^{d_2\times N}
~\text{and}~\widetilde{\overline{\textbf{X}}}_{\textbf{I}}=\begin{pmatrix}\textbf{w}^T\overline{\textbf{X}}_{t_1}\\ \vdots\\\textbf{w}^T\overline{\textbf{X}}_{t_N}\end{pmatrix}^T\in\mathbb{R}^{d_2\times N}.
\end{equation}
Then $\widetilde{\boldsymbol{\Sigma}}=(\textbf{D}-\widetilde{\overline{\textbf{X}}}_{\textbf{I}})
(\textbf{D}-\widetilde{\overline{\textbf{X}}}_{\textbf{I}})^T$.

According to \cite{Fukunaga90}, we have
\begin{equation}\label{Bbound}
\begin{split}
\int\sqrt{p_i(\widetilde{\textbf{X}})p_{j}(\widetilde{\textbf{X}})}= e^{-\frac{1}{8}(\widetilde{{\overline{\textbf{X}}}_{i}}-\widetilde{{\overline{\textbf{X}}}_{j}})\widetilde{\boldsymbol{\Sigma}}^{-1}(\widetilde{{\overline{\textbf{X}}}_{i}}-\widetilde{{\overline{\textbf{X}}}_{j}})^T}.
\end{split}
\end{equation}

The upper bound of the error $\epsilon_B$ can be estimated as
\begin{equation}\label{Bhattacharyya1}
\begin{split}
\epsilon_B
&=\sum\limits_{i<j}^c \sqrt{P_iP_j} e^{-\frac{1}{8}(\widetilde{{\overline{\textbf{X}}}_{i}}-\widetilde{{\overline{\textbf{X}}}_{j}})\widetilde{\boldsymbol{\Sigma}}^{-1}(\widetilde{{\overline{\textbf{X}}}_{i}}-\widetilde{{\overline{\textbf{X}}}_{j}})^T}\\
&=\sum\limits_{i<j}^c \sqrt{P_iP_j} e^{-\frac{1}{8}||(\widetilde{{\overline{\textbf{X}}}_{i}}-\widetilde{{\overline{\textbf{X}}}_{j}})\widetilde{\boldsymbol{\Sigma}}^{-\frac{1}{2}}||_2^2}\\
&\leq\sum\limits_{i<j}^c \sqrt{P_iP_j} \left(1-\frac{a}{8}||(\widetilde{{\overline{\textbf{X}}}_{i}}-\widetilde{{\overline{\textbf{X}}}_{j}})\widetilde{\boldsymbol{\Sigma}}^{-\frac{1}{2}}||_2^2\right)\\
&=\sum\limits_{i<j}^c \sqrt{P_iP_j} -\frac{a}{8}\sum\limits_{i<j}^c\sqrt{P_iP_j}\cdot ||(\textbf{w}^T{{\overline{\textbf{X}}}_{i}}-\textbf{w}^T{{\overline{\textbf{X}}}_{j}})\widetilde{\boldsymbol{\Sigma}}^{-\frac{1}{2}}||_2^2\\
&\leq\sum\limits_{i<j}^c \sqrt{P_iP_j} -\frac{a}{8}\sum\limits_{i<j}^c\sqrt{P_iP_j}\cdot \frac{||(\textbf{w}^T{{\overline{\textbf{X}}}_{i}}-\textbf{w}^T{{\overline{\textbf{X}}}_{j}})||_2^2}
{||\widetilde{\boldsymbol{\Sigma}}^{\frac{1}{2}}||_F^2}\\
&\leq\sum\limits_{i<j}^c \sqrt{P_iP_j} -\frac{a}{8}\sum\limits_{i<j}^c\sqrt{P_iP_j}\cdot ||\textbf{w}^T({\overline{\textbf{X}}_{i}-\overline{\textbf{X}}_{j}})||_2^2\\
&~~~+\frac{a}{8}\sum\limits_{i<j}^c\sqrt{P_iP_j}\cdot \Delta_{ij}'||\widetilde{\boldsymbol{\Sigma}}^{\frac{1}{2}}||_F^2,
\end{split}
\end{equation}
where $\Delta_{ij}'= \frac{1}{4}||{\overline{\textbf{X}}_{i}-\overline{\textbf{X}}_{j}}||_F^2$, $a>0$ is some constant.
For the first inequality of \eqref{Bhattacharyya1}, note that the real value function $f(z)=e^{-z}$ is concave when $z\in[0,b]$, therefore $e^{-z}\leq 1-\frac{1-e^{-b}}{b}z$. By taking $a=\frac{1-e^{-b}}{b}$ and noting $\widetilde{{\overline{\textbf{X}}}_{i}}=\textbf{w}^T{\overline{\textbf{X}}}_{i}$, the first inequality is obtained.
For the second inequality, we first note the fact that for any $\textbf{z}\in\mathbb{R}^{1\times {d_2}}$ and an invertible $\textbf{A}\in\mathbb{R}^{d_2\times d_2}$, $||\textbf{z}||_2=||(\textbf{z}\textbf{A})\textbf{A}^{-1}||_2\leq
||\textbf{z}\textbf{A}||_2\cdot||\textbf{A}^{-1}||_F$, which implies $||\textbf{z}\textbf{A}||_2\geq \frac{||\textbf{z}||_2}{||\textbf{A}^{-1}||_F}$. By taking $\textbf{z}=\textbf{w}^T{{\overline{\textbf{X}}}_{i}}-\textbf{w}^T{{\overline{\textbf{X}}}_{j}}$ and $\textbf{A}=\widetilde{\boldsymbol{\Sigma}}^{-\frac{1}{2}}$, we get the second inequality.
For the last inequality, since $||\textbf{w}||_2=1$, $||\textbf{w}^T({\overline{\textbf{X}}_{i}-\overline{\textbf{X}}_{j}})||_2^2\leq||\textbf{w}||_2^2\cdot||{\overline{\textbf{X}}_{i}-\overline{\textbf{X}}_{j}}||_F^2 = ||{\overline{\textbf{X}}_{i}-\overline{\textbf{X}}_{j}}||_F^2$ and $\frac{1}{||\widetilde{\boldsymbol{\Sigma}}^{\frac{1}{2}}||_F^2}
\left(1-\frac{1}{||\widetilde{\boldsymbol{\Sigma}}^{\frac{1}{2}}||_F^2}\right)\leq \frac{1}{4}$, we have
\begin{equation}\label{ueq1}
\begin{split}
&\left(||\textbf{w}^T({\overline{\textbf{X}}_{i}-\overline{\textbf{X}}_{j}})||_2^2
-\frac{||\textbf{w}^T({\overline{\textbf{X}}_{i}-\overline{\textbf{X}}_{j}})||_2^2}
{||\widetilde{\boldsymbol{\Sigma}}^{\frac{1}{2}}||_F^2}\right)\cdot\frac{1}{||\widetilde{\boldsymbol{\Sigma}}^{\frac{1}{2}}||_F^2}\\
=&||\textbf{w}^T({\overline{\textbf{X}}_{i}-\overline{\textbf{X}}_{j}})||_2^2\cdot\frac{1}{||\widetilde{\boldsymbol{\Sigma}}^{\frac{1}{2}}||_F^2}
\left(1-\frac{1}{||\widetilde{\boldsymbol{\Sigma}}^{\frac{1}{2}}||_F^2}\right)\\
\leq &\frac{1}{4}||\textbf{w}^T({\overline{\textbf{X}}_{i}-\overline{\textbf{X}}_{j}})||_2^2\\
\leq &\frac{1}{4}||{\overline{\textbf{X}}_{i}-\overline{\textbf{X}}_{j}}||_F^2\\
= &\Delta_{ij}'.
\end{split}
\end{equation}
which implies
\begin{equation}\label{ueq2}
\begin{split}
&-\frac{||\textbf{w}^T({\overline{\textbf{X}}_{i}-\overline{\textbf{X}}_{j}})||_2^2}
{||\widetilde{\boldsymbol{\Sigma}}^{\frac{1}{2}}||_F^2}
\leq -||\textbf{w}^T({\overline{\textbf{X}}_{i}-\overline{\textbf{X}}_{j}})||_2^2
+\Delta_{ij}'\cdot ||\widetilde{\boldsymbol{\Sigma}}^{\frac{1}{2}}||_F^2.
\end{split}
\end{equation}
By multiplying $\frac{a}{8}\sqrt{P_iP_j}$ to both sides of \eqref{ueq2} and summing it over all $1\leq i<j\leq c$, we get the last inequality of \eqref{Bhattacharyya1}.

Take $\Delta=\sum\limits_{i<j}^c\sqrt{P_iP_j}\Delta_{ij}'= \frac{1}{4}\sum\limits_{i<j}^c\sqrt{P_iP_j}||{\overline{\textbf{X}}_{i}-\overline{\textbf{X}}_{j}}||_F^2$, and note that $||\widetilde{\boldsymbol{\Sigma}}^{\frac{1}{2}}||_F^2=\sum_{i=1}^{c}\sum_{s=1}^{N_i}||\textbf{w}^T(\textbf{X}_{is}-\overline{\textbf{X}}_i)||_2^2$,
we then obtain \eqref{BhattacharyyaL2}. \hfill$\square$

\section*{Acknowledgment}

This work is supported by the National Natural Science Foundation of China (No.61703370, No.11871183, No.61866010, No.11771275 and No.61603338), in part Zhejiang Provincial Natural Science Foundation (No.LQ17F030003 and No.LY18G010018), and in part by Natural Science Foundation of Inner Mongolia Autonomous Region (No.2019BS01009).


%
%
%


\begin{thebibliography}{}

\end{thebibliography}


\begin{thebibliography}{99}
\bibitem{Fisher36}
Fisher R A. The use of multiple measurements in taxonomic problems. Annals of Eugenics, 1936, 7(2): 179-188.

\bibitem{Fukunaga90}
Fukunaga K. Introduction to statistical pattern recognition, second edition. Academic Press, New York, 1991.

\bibitem{BelhumeurHespanha97}
Belhumeur P N, Hespanha J P, Kriegman D J, et al. Eigenfaces vs. Fisherfaces: recognition using class specific linear projection. IEEE Transactions on Pattern Analysis and Machine Intelligence, 1997, 19(7): 711-720.

\bibitem{AbuZeina18}
AbuZeina D, Al-Anzi F S. Employing fisher discriminant analysis for Arabic text classification. Computers \& Electrical Engineering, 2018, 66: 474-486.

\bibitem{Zeiler16}
Zeiler S, Nicheli R, Ma N, et al. Robust audiovisual speech recognition using noise-adaptive linear discriminant analysis. 2016 IEEE International Conference on Acoustics, Speech and Signal Processing (ICASSP), 2016: 2797-2801.

\bibitem{GuoHastie07}
Guo Y, Hastie T, Tibshirani R. Regularized linear discriminant analysis and its application in microarrays. Biostatistics, 2007, 8(1): 86-100.

\bibitem{DongZhao16}
Dong K, Zhao H, Tong T, et al. NBLDA: negative binomial linear discriminant analysis for RNA-Seq data. BMC bioinformatics, 2016, 17(1): 369.

\bibitem{LiPangYuan10}
Li X, Pang Y, Yuan Y. L1-norm-based 2DPCA. Systems, Man, and Cybernetics, Part B: IEEE Transactions on Cybernetics, 2010, 40(4): 1170-1175.

\bibitem{Liuetal93}
Liu K, Cheng Y Q, Yang J Y. Algebraic feature extraction for image recognition based on an optimal discriminant criterion. Pattern Recognition, 1993, 26(6): 903-911.

\bibitem{LiYuan05}
Li M, Yuan B. 2D-LDA: A statistical linear discriminant analysis for image matrix. Pattern Recognition Letters, 2005, 26(5): 527-532.

\bibitem{XiongSwamy05}
Xiong H, Swamy M N S, Ahmad M O. Two-dimensional FLD for face recognition. Pattern Recognition, 2005, 38(7): 1121-1124.

\bibitem{Kong05}
Kong H, Teoh E K, Wang J G, et al. Two-dimensional Fisher discriminant analysis: forget about small sample size problem. IEEE International Conference on Acoustics, Speech, and Signal Processing, 2005: 761-764.

\bibitem{LiShangShao19}
Li C N, Shang M Q, Shao Y H, et al. Sparse L1-norm two dimensional linear discriminant analysis via the generalized elastic net regularization. Neurocomputing, 2019, 337: 80-96.

\bibitem{L12DLDA15}
Li C N, Shao Y H, Deng N Y. Robust L1-norm two-dimensional linear discriminant analysis. Neural Networks, 2015, 65: 92-104.

\bibitem{ChenL12DLDA15}
Chen S B, Chen D R, Luo B. L1-norm based two-dimensional linear discriminant analysis (In Chinese). Journal of Electronics and Information Technology, 2015, 37(6): 1372-1377.

\bibitem{TrL12DLDA17}
Li M, Wang J, Wang Q, et al. Trace ratio 2DLDA with L1-norm optimization. Neurocomputing, 2017, 266(29): 216-225.

\bibitem{LuYuanLai18}
Lu Y, Yuan C, Lai Z, et al. Horizontal and vertical nuclear norm-based 2DLDA for image representation. IEEE Transactions on Circuits and Systems for Video Technology, 2018, 29(4): 941-955.

\bibitem{ZhangDengNie19}
Zhang P, Deng S, Nie F, et al. Nuclear-norm based 2DLDA with application to face recognition. Neurocomputing, 2019, 339: 94-104.

\bibitem{G2DLDA}
Li C N, Shao Y H, Chen W J, et al. Generalized two-dimensional linear discriminant analysis with regularization. arXiv preprint arXiv:1801.07426, 2018.

\bibitem{LiShaoWang19}
Li C N, Shao Y H, Wang Z, et al. Robust bilateral Lp-norm two-dimensional linear discriminant analysis. Information Sciences, 2019, 500: 274-297.

\bibitem{Duetal17}
Du H, Zhao Z, Wang S, et al. Two-dimensional discriminant analysis based on Schatten p-norm for image feature extraction. Journal of Visual Communication and Image Representation, 2017, 45: 87-94.

\bibitem{YangZhang05}
Yang J, Zhang D, Yong X, Yang J Y. Two-dimensional discriminant transform for face recognition. Pattern Recognition, 2005, 38: 1125-1129.

\bibitem{Noushath06}
Noushath S, Kumar G H, Shivakumara P. (2D)$^2$LDA: An efficient approach for face recognition. Pattern recognition, 2006, 39(7): 1396-1400.

\bibitem{YeJanardan05}
Ye J, Janardan R, Li Q. Two-dimensional linear discriminant analysis. Advances in Neural Information Processing Systems. 2005: 1569-1576.

\bibitem{WangQinetal17}
Wang Q, Qin Z, Nie F, et al. Convolutional 2D LDA for nonlinear dimensionality reduction. international joint conference on artificial intelligence, 2017: 2929-2935.

\bibitem{XiaoChenGong19}
Xiao X, Chen Y, Gong Y J, et al. Two-dimensional quaternion sparse discriminant analysis. IEEE Transactions on Image Processing, 2019, 29: 2271-2286.

\bibitem{L2BLDA}
Li C N, Shao Y H, Wang Z, et al. Robust Bhattacharyya bound linear discriminant analysis through an adaptive algorithm. Knowledge-Based Systems, 2019, 183: 104858.

\bibitem{Bhattacharyya43}
Bhattacharyya A. On a measure of divergence between two statistical populations defined by their probability distribution. Bulletin of Calcutta Mathematical Society, 1943.

\bibitem{Devijveretal82}
Devijver P A, Kittler J. Pattern recognition: a statistical approach. Prentice/hall International, 1982.

\bibitem{Saon02}
Saon G, Padmanabhan M. Minimum Bayes error feature selection. Proceeding of NIPS, 2002: 800-806.

\bibitem{RuedaHerrera08}
Rueda L, Herrera M. Linear dimensionality reduction by maximizing the Chernoff distance in the transformed space. Pattern Recognition, 2008, 41(10): 3138-3152.

\bibitem{Nielsen14}
Nielsen F. Generalized Bhattacharyya and Chernoff upper bounds on Bayes error using quasi-arithmetic means. Pattern Recognition Letters, 2014, 42:25-34.

\bibitem{YangZhang04}
Yang J, Zhang D, Frangi A F, et al. Two-dimensional PCA: a new approach to appearance-based face representation and recognition. IEEE Transactions on Pattern Analysis and Machine Intelligence, 2004, 26(1): 131-137.

\bibitem{NeneNayar96}
Nene S A, Nayar S K, Murase H. Columbia Object Image Library (COIL-100), Technical Report CUCS-006-96, February 1996.
\end{thebibliography}
\end{document}